\documentclass{article}

\usepackage{graphicx}
\usepackage{subfigure}
\usepackage{booktabs} 
\usepackage{hyperref}
\usepackage{makecell}

\usepackage[accepted]{icml2023}
\usepackage{amsmath}
\usepackage{amssymb}
\usepackage{mathtools}
\usepackage{amsthm}
\usepackage{amsmath}
\usepackage{listings}
\usepackage[capitalize,noabbrev]{cleveref}
\usepackage{multirow}
\usepackage{pifont}
\usepackage{mathtools}
\usepackage{enumitem}
\usepackage{setspace}
\usepackage{soul}
\theoremstyle{plain}

\theoremstyle{definition}

\theoremstyle{remark}

\newcommand\our{{MH-MoE}}
\newcommand\ourfull{{Multi-Head Mixture-of-Experts}}
\newcommand\xmoe{{X-MoE}}
\newcommand\base{{Dense}}

\usepackage[textsize=tiny]{todonotes}
\newcommand{\red}[1]{\textcolor{red}{#1}}

\definecolor{gray}{rgb}{0.5,0.5,0.5}
\definecolor{darkergreen}{RGB}{21, 152, 56}
\definecolor{darkerblue}{rgb}{0,0.08,0.45}
\definecolor{RoyalBlue}{RGB}{65,105,225}
\definecolor{YellowOrange}{RGB}{255,165,0}
\definecolor{gray94}{gray}{.92}
\definecolor{gray90}{gray}{.90}
\definecolor{gray85}{gray}{.85}
\usepackage{xcolor}  
\usepackage{wrapfig}
\usepackage{colortbl}
\usepackage{pifont}
\usepackage{xcolor}
\usepackage{wrapfig}
\usepackage{epstopdf}
\usepackage{wrapfig}
\usepackage{cuted}
\usepackage{capt-of}
\usepackage{hyperref}

\newcommand{\cmark}{{\color{blue}\ding{51}}}%
\newcommand{\xmark}{{\color{red}\ding{55}}}%

\icmltitlerunning{\ourfull{}}

\begin{document}

\twocolumn[
\icmltitle{\ourfull{}}




\begin{icmlauthorlist}
\icmlauthor{Xun Wu}{yyy,comp}
\icmlauthor{Shaohan Huang}{comp}
\icmlauthor{Wenhui Wang}{comp}
\icmlauthor{Furu Wei}{comp}
\end{icmlauthorlist}

\icmlaffiliation{yyy}{Tsinghua University, Beijing, China}
\icmlaffiliation{comp}{Microsoft Research, Beijing, China}

\icmlcorrespondingauthor{Shaohan Huang}{shaohanh@microsoft.com}

\icmlkeywords{Machine Learning, ICML}

\vskip 0.3in
]
\printAffiliationsAndNotice{}



\begin{abstract}
Sparse Mixtures of Experts (SMoE) scales model capacity without significant increases in training and inference costs. However, it exhibits two issues: (1) \textit{Low expert activation}, where only a small subset of experts are activated for optimization, leading to suboptimal performance and limiting its effectiveness in learning a larger number of experts in complex tasks. (2) \textit{Lack of fine-grained analytical capabilities} for multiple semantic concepts within individual tokens. 
In this paper,~we propose \ourfull{} (\our{}). \our{} employs a multi-head mechanism to split each input token into multiple sub-tokens. Then these sub-tokens are assigned to and processed by a diverse set of experts in parallel, and seamlessly reintegrated into the original token form. 
The above operations enables \our{} to collectively attend to information from various representation spaces within different experts to deepen context understanding while significantly enhancing expert activation.
It's worth noting that our \our{} is straightforward to implement and decouples from other SMoE frameworks, making it easy to integrate with these frameworks for enhanced performance.
Extensive experimental results across three tasks: English-focused language modeling, Multi-lingual language modeling and Masked multi-modality modeling tasks, demonstrate the effectiveness of \our{}. Our code are available at \url{https://github.com/yushuiwx/MH-MoE} .
\end{abstract}

\section{Introduction}
Large capacity models, such as Large Language Models~(LLMs)~\citep{zhao2023sparse, pham2023task, chung2022scaling, OpenAI2023GPT4TR} and Large Multi-modal Models (LMMs)~\citep{wang2022BEiTv3,peng2023kosmos}, have demonstrated their efficacy across various domains and tasks.
\begin{figure}[t]
\centering
\raisebox{6\height}{\makebox[0.055\textwidth]{\makecell{\small (a)}}}
\includegraphics[width=0.8\linewidth]{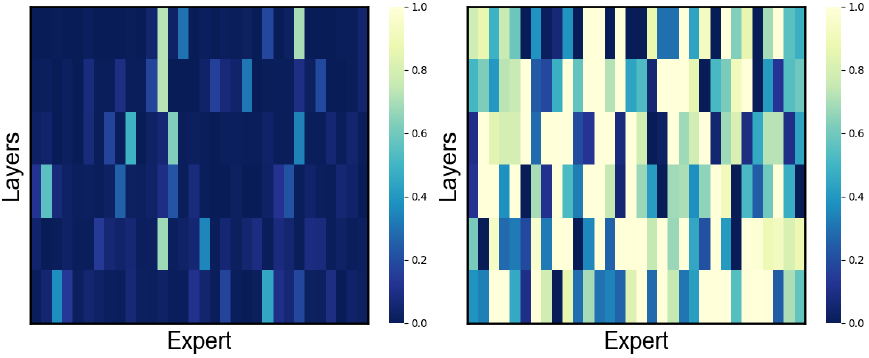}
\\
\vspace{-1mm}
\centering
\hspace{8mm}
\makebox[0.4\linewidth]{\small SMoE}
\makebox[0.4\linewidth]{\small \our{}}
\\
\vspace{-1.5mm}
\hspace{8mm}
\makebox[0.4\linewidth]{\scriptsize Activation: \red{8.33\%}}
\makebox[0.4\linewidth]{\scriptsize Activation: \red{90.71\%}}
\vspace{2mm}
\\
\centering
\raisebox{5\height}{\makebox[0.055\textwidth]{\makecell{\small (b)}}}
\includegraphics[width=0.78\linewidth]{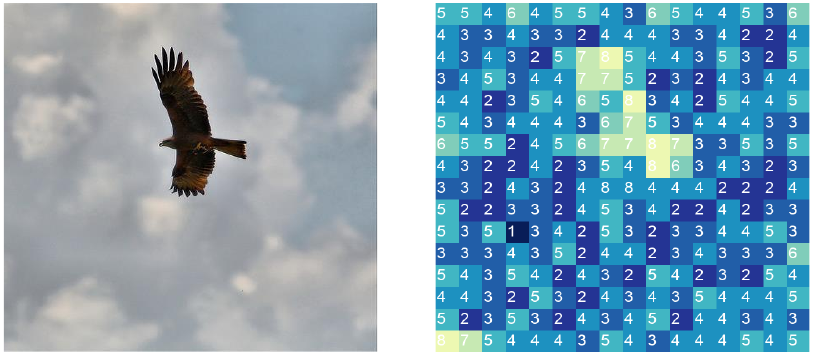}
\\
\vspace{-1mm}
\centering
\hspace{8mm}
\makebox[0.4\linewidth]{\small RGB Input}
\makebox[0.4\linewidth]{\footnotesize Sub-tokens Assign}
\\
\vspace{-2mm}
\caption{(a) \textbf{Expert activation distribution} on XNLI~\citep{xnli}, encompassing $6$ parallel expert layers with $32$ experts per layer. SMoE has many ``dead'' experts (dark) which are not activated, while \our{} leading to significantly increased usage of these experts. Experts activation ratio is determined by calculating the ratio of each expert's selection frequency in each MoE layer to the total number of tokens, where those exceeding a threshold ($<$1) are considered activated. (b) \textbf{\our{} showcases finer-grained understanding} by distributing sub-tokens split from semantically-rich patches to more distinct experts to capture semantic information. Brighter regions indicate that sub-tokens from this patch are distributed to a greater number of diverse experts, while darker regions indicate that sub-tokens are assigned to more of the same experts.}
\label{fig:motivation}
\vspace{-4mm}
\end{figure}
To further enhance performance, a reliable approach involves scaling up these models by augmenting the parameter count~\citep{fedus2022switch}. But for most of these densely-activated large-capacity models (referred to as Dense models), which utilize all their parameters to process all inputs, the extremely large size of these models significantly reduces inference speed, further limiting their practicality.

\begin{figure*}[t]
\centering
\includegraphics[width=\linewidth]{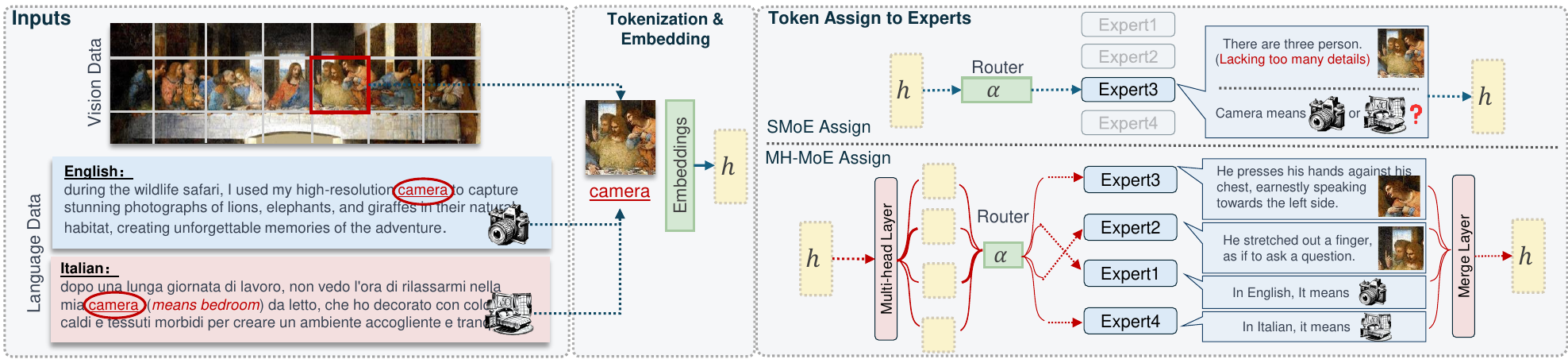}
\vspace{-7mm}
\caption{\textbf{Workflow for \our{} on both vision and language data}. For vision data, different heads routed to different experts try to capture different aspects of details within patches and relations between patches. For language data, different heads attend to capture the varying contexts of false cognates across different languages (e.g., Italian and English) or polysemous words within a single language.}
\label{fig:workflow}
\vspace{-4mm}
\end{figure*}

A promising alternative, facilitating model scalability while mitigating the burdensome computational costs, resides in Sparse Mixtures of Experts (SMoE)~\citep{smoe, du2021glam, chi2022representation, clark2022unified}. In contrast to Dense model, SMoE contains parallel feed-forward neural networks (referred to as experts) within each building block, and strategically activates distinct experts for specific input tokens via a router, thereby yielding noteworthy efficiency enhancements. 
%
GShard~\citep{lepikhin2020gshard} scales a Dense model from 2B to 600B parameters with lower training costs than a 100B Dense model. And recently, Mixtral 8$\times$7B~\citep{Mixtral}, a SMoE model containing 8 experts (7B parameter in total) is shown to outperform or matches LLaMA-2 70B~\citep{touvron2023llama} and \texttt{GPT-3.5}.

Despite its success, SMoE has some drawbacks: 
(1) \emph{Low experts activation}, which means that only a small subset of experts are activated during optimization and inference, e.g., \textbf{8.33\%} activation ratio shown in Figure~\ref{fig:motivation}~(a), while the majority of them are not used at all (see the dark area). As a result, SMoE fails to utilize the full expressive power of these experts, especially when the number of experts is large, which significantly limits effectiveness and scalability of SMoE. 
%
%
(2) \emph{Absence of fine-grained analytical capabilities.} The current tokenization patterns impose limitations on the model's capacity to grasp multiple semantic interpretations linked to individual tokens. In the context of visual data, dividing images into patches for tokenization may either neglect finer image details when using larger patches or escalate computational requirements when employing smaller ones. For language data, the tokenization of false cognates across different languages or polysemous words within a single language results in them being represented by the same tokens, despite carrying distinct meanings. This can subsequently lead to confusion within the models.

%
To tackle the above issues, we propose Multi-Head Mixture-of-Experts~(\our). The workflow of \our{} is illustrated in Figure~\ref{fig:workflow}. By employing a multi-head mechanism to split each input token into multiple sub-tokens and distribute them to different experts, \our{} achieves denser expert activation without an increase in computational and parameter complexity. Specifically, as shown in Figure~\ref{fig:workflow}, when provided with a single input token, \our{} activates four experts by splitting it into four sub-tokens, whereas SMoE only activates one expert. 
Furthermore, the allocation of sub-tokens to distinct experts enables the model to simultaneously focus on information from various representation spaces within different experts, ensuring a more granular understanding for subtle differences in both vision and language patterns. See in Figure~\ref{fig:workflow}, sub-tokens assigned to Experts 3 and 2 capture a detailed understanding of each character's actions within an image patch, while those assigned to Experts 1 and 4 explicitly model the semantics of the false cognate `camera'.
After expert processing, sub-tokens are seamlessly reintegrated into the original token form, thereby circumventing any additional computational burden in subsequent non-parallel layers, e.g., attention layer, while also integrating semantic information captured from multiple experts
%

\our{} maintains following strengths: 
(1) \textbf{Higher experts activation \& better scalability}. \our{} can alleviate lower expert activation problem and significantly enhance the usage of larger experts by enabling optimization of almost all of experts, e.g., achieving \textbf{90.71\%} activation in Figure~\ref{fig:motivation} (a), allowing for more efficient scaling of model capacity. 
(2) \textbf{Finer-grained understanding ability}. Multi-head mechanism adopted in \our{} assign sub-tokens to different experts, enabling to jointly attend to information from different representation spaces at different experts, and finally achieving better finer-grained understanding ability. For example, refer to the bright area in Figure~\ref{fig:motivation}~(b), where sub-tokens are distributed among a more diverse set of experts, facilitating the capture of semantically-rich information.
(3) \textbf{Seamless integration}. The implementation of \our{} is remarkably straightforward and decoupled from other SMoE optimization methods~(e.g., GShard~\citep{lepikhin2020gshard}), making it very easy to integrate them together to achieve better performance.

We evaluate the proposed \our{} on three model pre-training and fine-tuning setting: English-focused language modeling, Multi-lingual language modeling and Masked multi-modality modeling. Extensive experimental among these three tasks demonstrate the effectiveness of \our{}.
%

\section{Background}
\label{Sec: Related Work}
\noindent\textbf{Sparse Mixture of Experts.} Sparse Mixture-of-Experts (SMoE)~\citep{smoe,du2021glam,chi2022representation, clark2022unified} enhances model capacity while maintaining a constant computational demand, thus achieving better performance than densely-activated models on various tasks~\citep{gshard, kumatani2021building, zhao2023sparse, pham2023task} and being emerged as a pivotal advancement in the field of deep learning.
%

Different from densely-activated models, each MoE layer consists of $N$ independent Feed-Forward Networks~(FFN) $\{f^{\text{FFN}}_{i}\}^{N}_{i=0}$ as the experts, along with a gating function $g \left(\cdot\right)$ to model a probability distribution indicating the weights over these experts' outputs.
For the hidden representation $\mathbf{h} \in \mathbb{R}^{d}$ of each input token, the gating value of routing $\mathbf{h}$ to expert $f^{\text{FFN}}_{i}$ is denoted as:
\begin{equation}
g \left(f^{\text{FFN}}_{i}\right) = \exp\left(\mathbf{h}\cdot\mathbf{e}_i\right) / \sum_{j=0}^{N}\exp\left(\mathbf{h}\cdot\mathbf{e}_j\right),
\label{Eq: Softmax in Routing}
\end{equation}
where $\mathbf{e}_i$ denotes the trainable embedding of the $i$-th expert and $\sum_{i=0}^{N}g \left(f^{\text{FFN}}_{i}\right) = 1$. Then, the corresponding $k$ experts, according to the top-$k$ gated values, are activated and the output $\mathbf{O}$ of the MoE layer is
\begin{equation}
\mathbf{O} = \mathbf{h} + \sum_{i \in \Phi}g \left(f^{\text{FFN}}_{i}\right) \cdot f^{\text{FFN}}_{i}\left(\mathbf{h}\right).
\end{equation}
where $\Phi$ denote activated experts set and $|\Phi| = k$.
\begin{figure*}[t]
\centering
\includegraphics[width=0.35\textwidth]{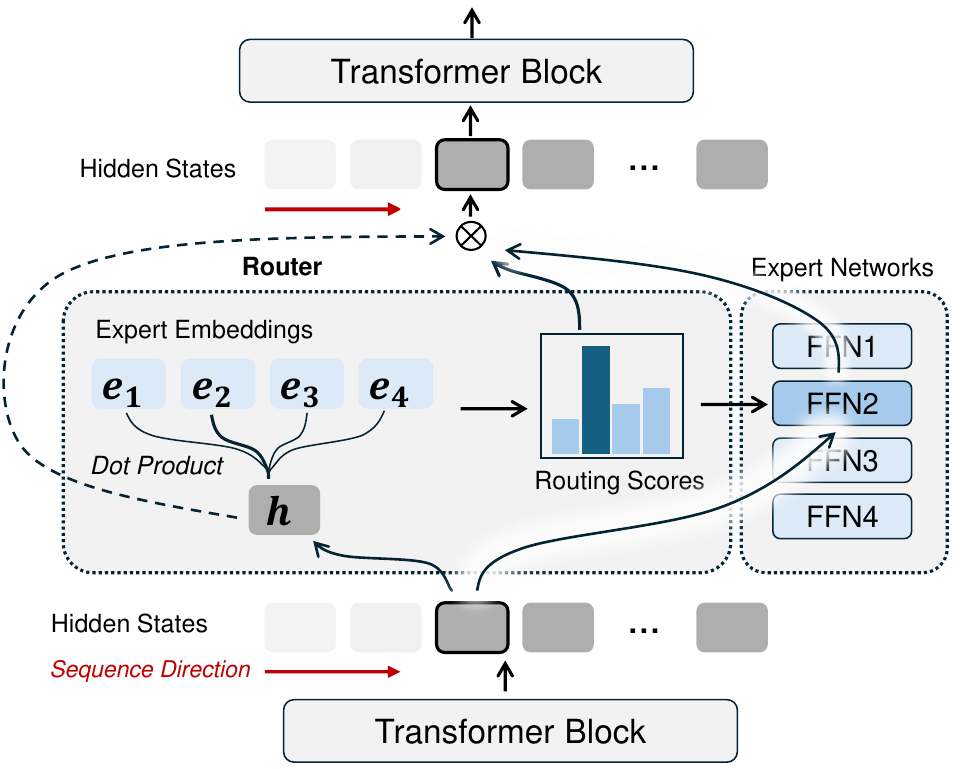}
\hspace{9mm}
\includegraphics[width=0.35\textwidth]{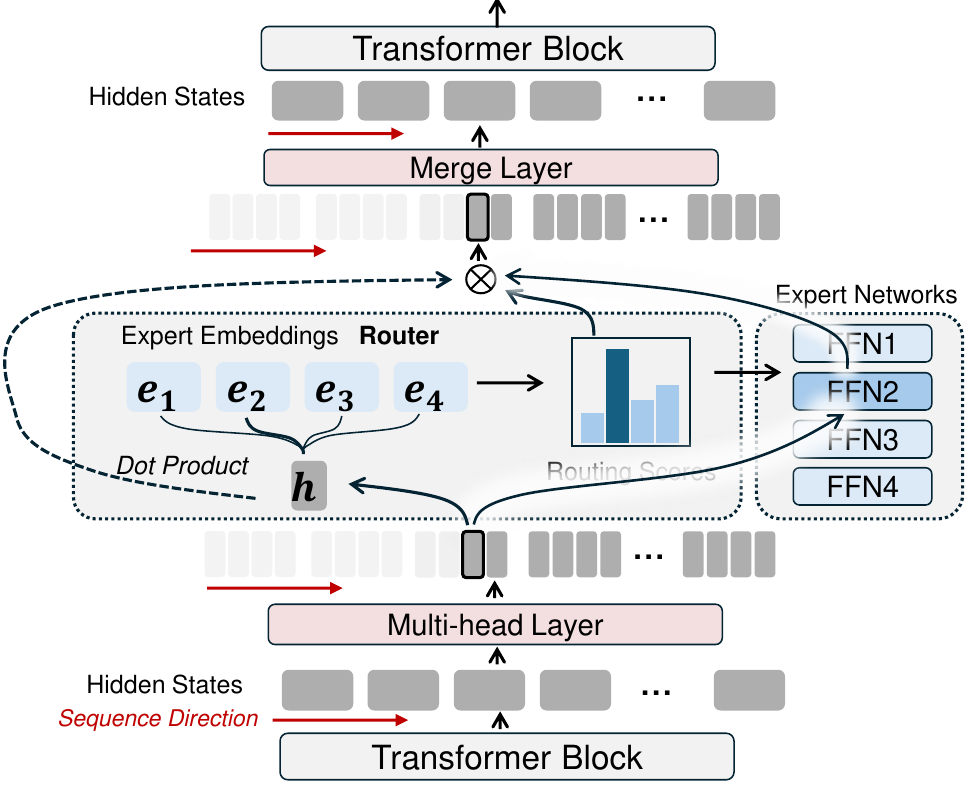}
 \\
\makebox[0.4\textwidth]{\small (a) SMoE}
\makebox[0.4\textwidth]{\small (b) \our{}}
\vspace{-3mm}
\caption{\textbf{Illustration of a typical SMoE layer and the proposed \our{} layer}. (a) An SMoE layer consists of a router and expert networks, where the experts are sparsely activated according to dot-product token-expert routing scores. (b) \our{} introduces additional two MLP layers, namely the multi-head layer and merge layer, and a Token-Splitting-Merging (TSM, Eq.~\ref{EQ. sub-tokens} and Eq.~\ref{EQ: merge tokens}) operation incorporated between these two MLPs.}
\label{fig:Main_arch}
\vspace{-4mm}
\end{figure*}

\noindent\textbf{Routing Mechanism in SMoE.} As described above, the most commonly used routing mechanism involves selecting the top-$k$ experts from $N$ experts, where $k \ll N$~\citep{shazeer2017outrageously},~e.g., $k$ = 2 and $N$ = 2048 in GShard~\citep{lepikhin2020gshard}. Such a routing mechanism allows the combination of
data parallelism and expert parallelism. \citet{yang2021m6} and \citet{lepikhin2020gshard} suggest that larger values of $k$ often contribute to better model performance. However, with the increase in the value of $k$, training models with conventional top-$k$ routing implementation becomes much less efficient~\citep{lepikhin2020gshard}. 

In this paper, we introduce \our{}, a simple but efficient manner to make denser expert activation without an increase in computational complexity.

\section{Method}
\label{Sec: Method}
The full architecture of \our{} can be seen in Figure~\ref{fig:Main_arch}, \our{} addresses low experts activation and confusion over ambiguity of tokens issues by applying a multi-head mechanism to split each token into sub-tokens and route them to various experts to achieve denser expert activation as well as deeper understanding.
%

\subsection{\ourfull{}}
\label{Sec:Routing Mechanism}
Concretely, we denote a sequence of inputs tokens by $\mathbf{X} \in \mathbb{R}^{l \times d}$, where $l$ is the number of tokens and $d$ represents the length of token dimension. In~\our{}, each parallel layer contains a set of $N$ experts, each presented as \{$f^{\text{FFN}}_{i}: \mathbb{R}^{\frac{d}{h}} \rightarrow \mathbb{R}^{\frac{d}{h}}$\}$^{N}_{i = 0}$, $h$ denotes the number of heads in multi-head mechanism, which is decoupled from the head in the multi-head self-attention layer. For clarity, we describe the operation of a single \our{} layer here only.

First, $\mathbf{X}$ is projected by a multi-head layer with parameter matrices $\mathbf{W}_{\text{head}} \in \mathbb{R}^{d \times d}$,
\begin{equation}
    \hat{\mathbf{X}} = \mathbf{X} \cdot \mathbf{W}^{\top}_{\text{head}}
    \label{EQ: multi-head}
\end{equation}
where $\hat{\mathbf{X}} \in \mathbb{R}^{l \times d}$. After that, every token in $\hat{\mathbf{X}}$ is split into $h$ sub-tokens along the token dimensions, and these sub-tokens are arranged in parallel according to the original token sequence, forming a new feature space $\ddot{\mathbf{X}} \in \mathbb{R}^{(l\times h) \times \frac{d}{h}}$ as:
\begin{align}
    \ddot{\mathbf{X}} &=  \digamma_{\text{s}} (\hat{\mathbf{X}}) \nonumber \\
    &= \left[\underbrace{\overbrace{\mathbf{x}^{0}_0,\ldots, \mathbf{x}^{0}_{h-1}}^{h}, \ldots ,\mathbf{x}^{i}_{j},\mathbf{x}^{i}_{j+1},\ldots,\mathbf{x}^{l}_{h-1}}_{l \times h}\right],
\label{EQ. sub-tokens}
\end{align}
where function $\digamma_{\text{s}}$ denotes the token splitting operation: $\mathbb{R}^{l \times d} \rightarrow \mathbb{R}^{\left(l \times h\right) \times \frac{d}{h}}$, and each sub-token is presented as $\mathbf{x}^{i}_{j} \in \mathbb{R}^{\frac{d}{h}}$, meaning it is the the $j^{th}$ sub-token split from the $i^{th}$ token. 

Then all these sub-tokens are fed into the gating function $g(\cdot)$. The gating value of routing a certain sub-token $\mathbf{x}^{i}_{j}$ into the $p^{th}$ expert is computed as
\begin{equation}
g \left(f^{\text{FFN}}_{p}\right) = \frac{\exp\left(\mathbf{x}^{i}_{j}\cdot\mathbf{e}_{p}\right)}{\sum_{\xi=0}^{N}\exp\left(\mathbf{x}^{i}_{j}\cdot\mathbf{e}_{\xi}\right)},
\label{Eq: routing score}
\end{equation}
where $\mathbf{e}_{p} \in \mathbb{R}^{\frac{d}{h}}$ is the learnable embedding of the $p^{th}$ expert.
In this paper, we mainly focus on top-$k$ routing,~\emph{i.e.}, only the experts with the largest top-$k$ routing score is activated.
$\Phi = \text{Top}_{k}\left(g \left(f^{\text{FFN}}\right)\right)$ denote the set of activated experts and $|\Phi| = k$. Then $\mathbf{x}^{i}_{j}$ is processed by these activated experts as following,
\begin{equation}
\mathbf{o}^{i}_{j} = \mathbf{x}^{i}_{j} + \sum_{p \in \Phi} g\left(f^{\text{FFN}}_{p}\right) \cdot f^{\text{FFN}}_{p}\left(\mathbf{x}^{i}_{j}\right).
\end{equation}

After that, all obtained $\mathbf{o}^{i}_{j}$ are rearranged in the original order of sub-tokens and integrated together as
\begin{equation}
    \mathbf{O} = \left[\underbrace{\overbrace{\mathbf{o}^{0}_0, \ldots \mathbf{o}^{0}_{h-1}}^{h}, \ldots, \mathbf{o}^{i}_{j},\mathbf{o}^{i}_{j+1},\ldots,\mathbf{o}_{h-1}^{l}}_{l \times h}\right],
    \label{EQ: merge tokens}
\end{equation}
where $\mathbf{O} \in \mathbb{R}^{\left(l \times h\right) \times \frac{d}{h}}$. 
After that, $\mathbf{O}$ is transformed back the into original token form by a token merging operation $\digamma_{\text{m}}$: $\mathbb{R}^{\left(l \times h\right) \times \frac{d}{h}} \rightarrow \mathbb{R}^{l \times d}$:
\begin{align}
    \bar{\mathbf{X}} &= \digamma_{\text{m}}\left(\mathbf{O}\right)^{\top},
    \label{EQ: merge tokens}
\end{align}

where $\bar{\mathbf{X}} \in \mathbb{R}^{l \times d}$. Finally, $\bar{\mathbf{X}}$ is projected by a merge layer with parameter matrices $\mathbf{W}_{\text{merge}} \in \mathbb{R}^{d \times d}$ to effective integration of multiple features $\mathbf{o}^{i}_{j}$ capturing detailed information from different expert representation spaces. The operation is presented as following:
\begin{align}
    \breve{\mathbf{X}} &= \bar{\mathbf{X}} \cdot\mathbf{W}_{\text{merge}}^{\top}.
    \label{EQ: merge layer}
\end{align}
Then we get the final output $\breve{\mathbf{X}}$ of the single MH-MoE layer.



We name the token splitting (Eq.~\ref{EQ. sub-tokens}) and token merging (Eq.~\ref{EQ: merge tokens}) operations together as the Token-Splitting-Mergin (TSM) operation. By implementing the aforementioned operations, we have effectively increased the average volume of data routed to a specific expert by a factor of $h$, as demonstrated in Eq.~\ref{EQ. sub-tokens}. Consequently, this achievement has resulted in denser expert activation. Furthermore, the allocation of sub-tokens to distinct experts within \our{} enables us to collectively capture semantic information from diverse feature spaces across these experts, thereby enhancing the model's ability to achieve a finer-grained understanding.

The operations mentioned above ensure that the shapes of the input and output in the \our{} layer remain unchanged. Consequently, no additional computational cost is introduced in the subsequent block. Specifically, we introduce a hyperparameter $\beta$ to scale the inner dimensions of each expert, aiming to balance the parameters introduced by the multi-head layer and merge layer, aligning the model's parameters and computational complexity with the original SMoE. 

As the Pytorch-like style pseudocode of~\our{} shown in Appendix~\ref{app: Algorithm}, \our{} is characterized by its overall simplicity of implementation, necessitating minimal modifications to the SMoE implementation. Additionally, it is decoupled from other SMoE optimization strategies~\citep{lepikhin2020gshard,chi2022representation}, thereby facilitating its convenient integration with other optimized SMoE frameworks to enhance performance.

\subsection{Training Objectives}
\label{Sec:Optimizing Objectives}
The training objective of \our{} involves the simultaneous minimization of both the loss associated with the target task and an auxiliary load balancing loss.

\noindent\textbf{Load balancing loss.} As described in Section~\ref{Sec: Related Work}, there is usually an expert load imbalance problem~\citep{xie2023moec,lepikhin2020gshard}. So, following~\citep{lepikhin2020gshard, fedus2022switch}, given the sub-token set $\ddot{\mathbf{X}}$ (depicted in Eq.~\ref{EQ. sub-tokens}) and the frequency $t_p$ of how many sub-tokens are routed to the $p^{th}$ expert, we compute the load balancing loss $\mathcal{L}_{\text{balance}}$ via:
%

\begin{equation}
 \mathcal{L}_{\text{balance}} = \frac{N}{|\ddot{\mathbf{X}}|} \sum_{p=1}^{N}\sum_{\mathbf{x}^{i}_{j} \in \ddot{\mathbf{X}}} t_p \cdot g \left(f^{\text{FFN}}_{p}\right),
\end{equation}

where $N$ denotes the number of experts, $|\ddot{\mathbf{X}}|$ is the number of sub-tokens contained in $\ddot{\mathbf{X}}$. $g \left(f^{\text{FFN}}_{p}\right)$ is the gating function depicted in Eq.~\ref{Eq: routing score}, denoting the gating value of routing a certain sub-token $\mathbf{x}^{i}_{j}$ into the $p^{th}$ expert.

\begin{figure*}[t]
    \centering
    \includegraphics[width=0.33\textwidth]{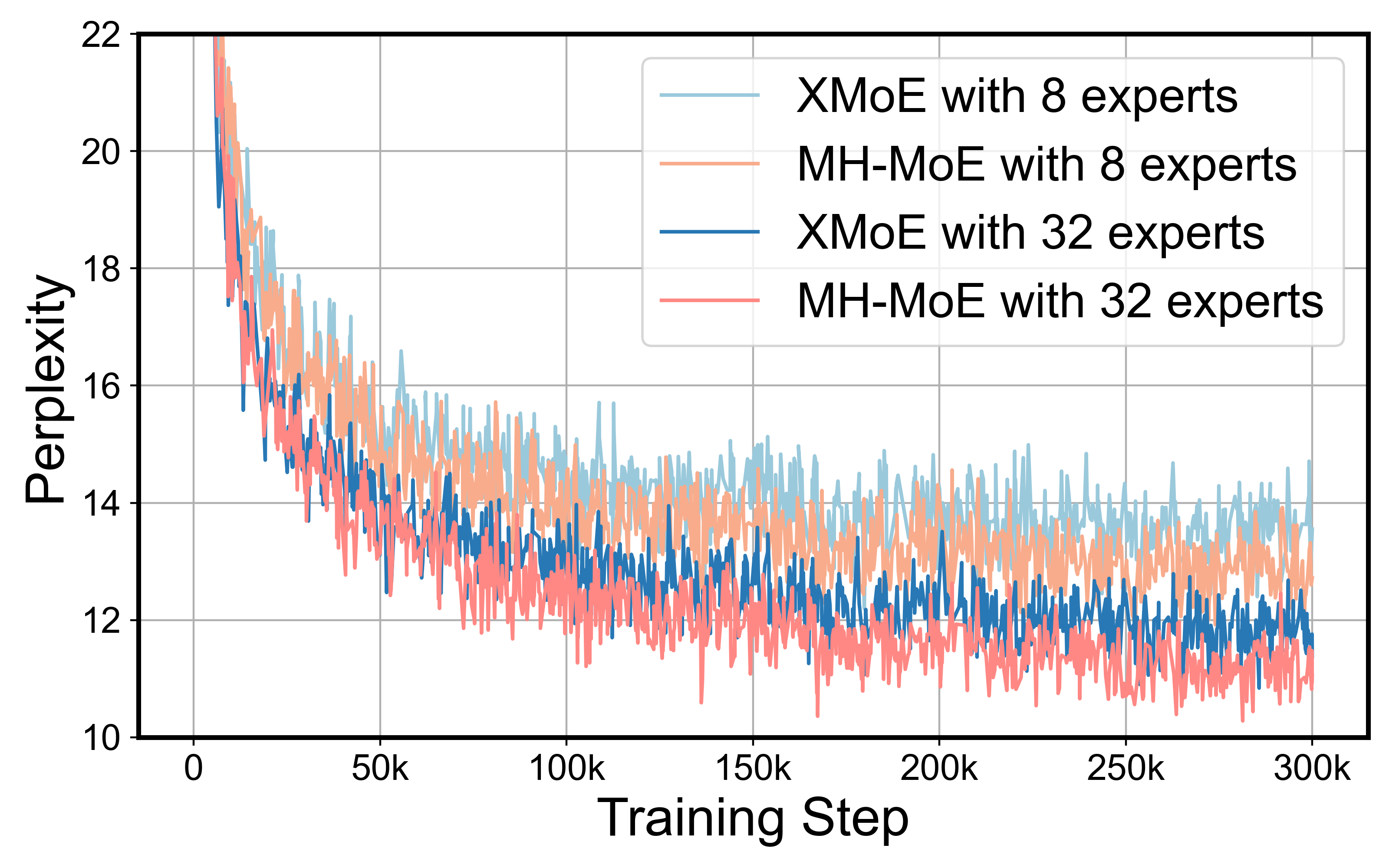}
    \includegraphics[width=0.33\textwidth]{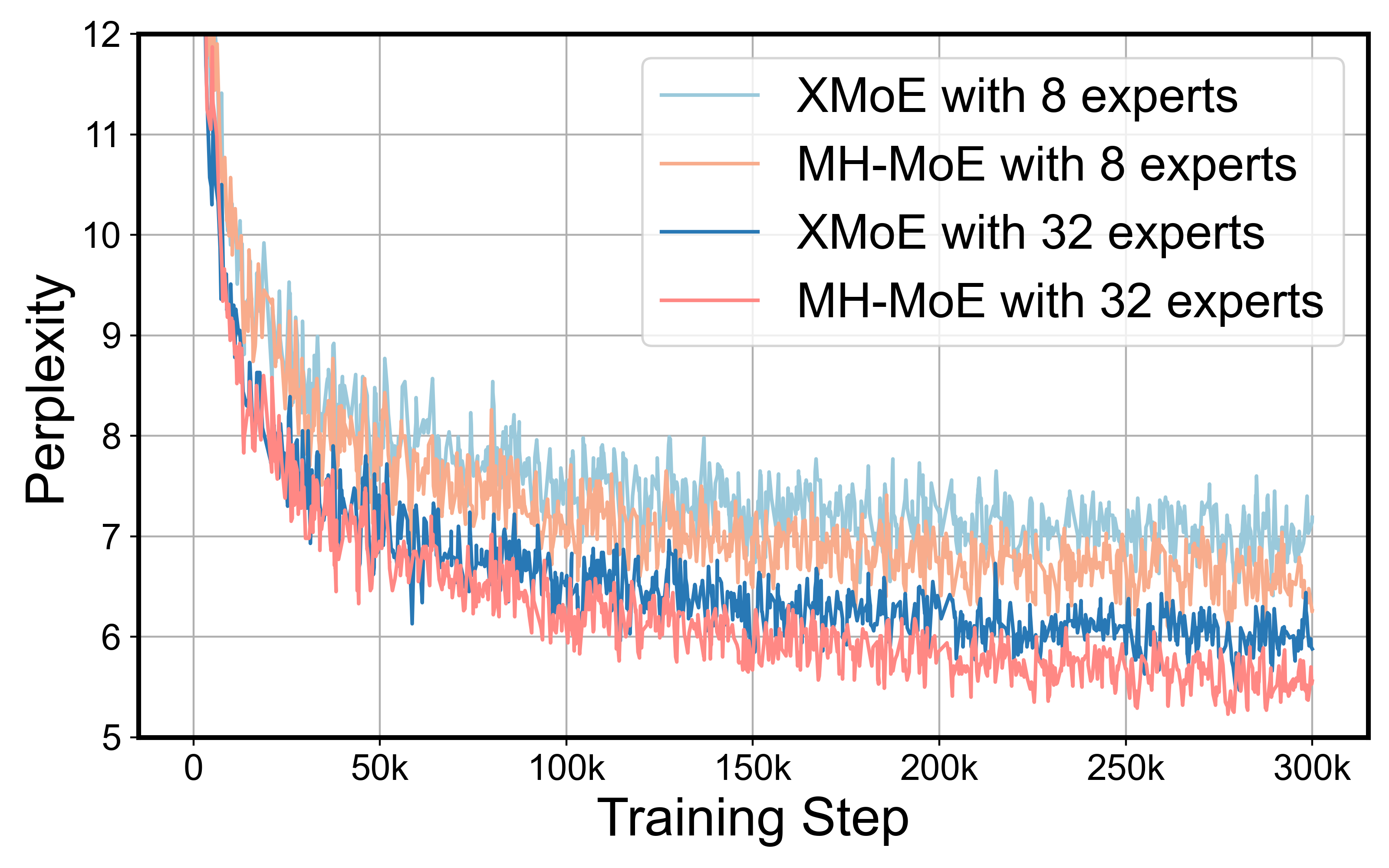}
    \includegraphics[width=0.33\textwidth]{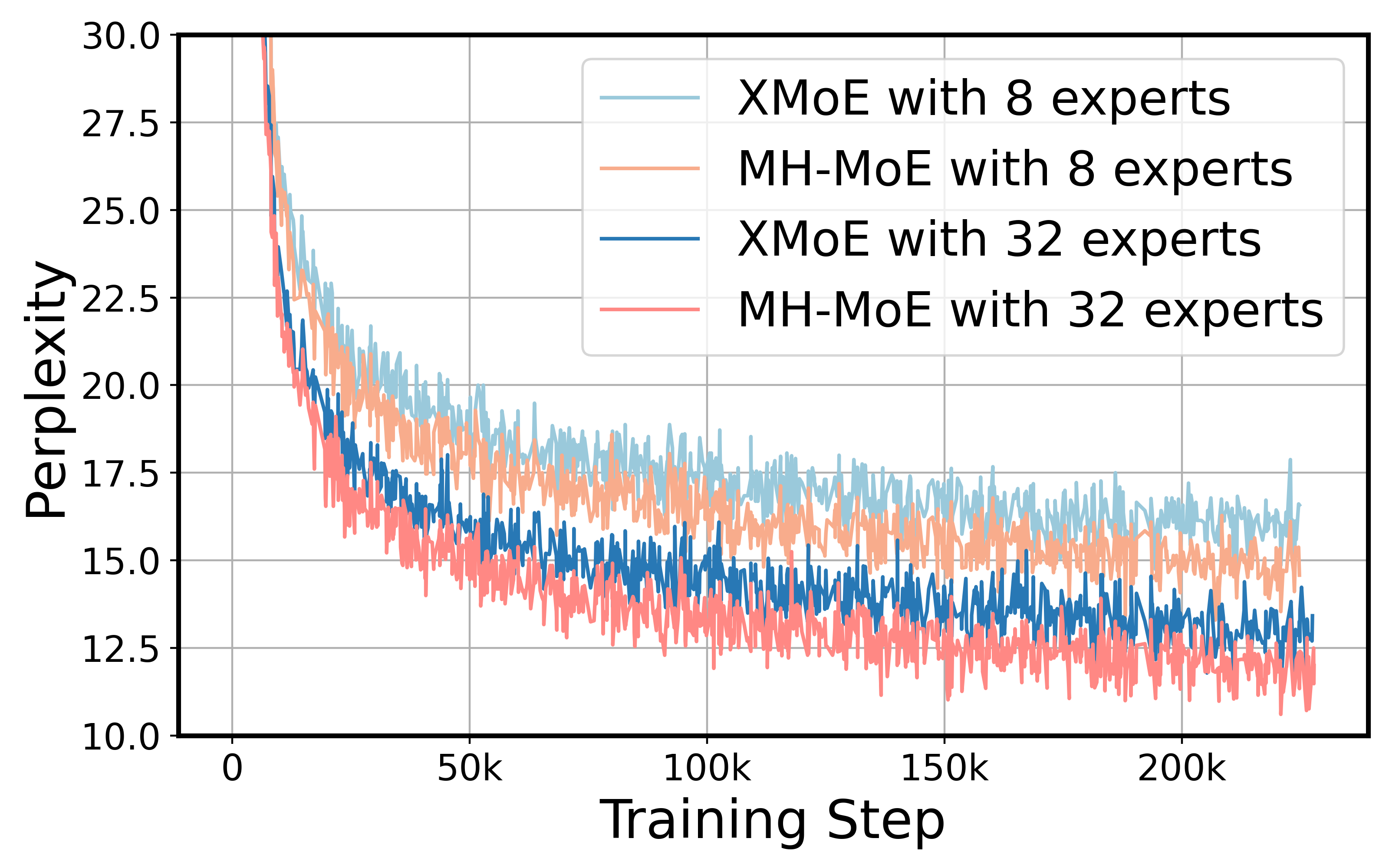}
    \\
    \vspace{-2mm}
    \makebox[0.33\textwidth]{\small (a)}
    \makebox[0.33\textwidth]{\small (b)}
    \makebox[0.33\textwidth]{\small (c)}
    \vspace{-8mm}
    \caption{\textbf{Perplexity on validation dataset during the training phase} reported for Dense,~\xmoe{} and \our{} across three pre-training tasks. (a) \textit{English-focused language modeling}. (b) \textit{Multi-lingual language modeling}. (c) \textit{Masked multi-modal modeling}}
    \label{fig:ppl}
    \vspace{-4mm}
\end{figure*}

\noindent\textbf{Task specific loss.} The term $\mathcal{L}_{\text{task}}$ is dependent on the particular task that \our~is designed to learn. For instance, during pre-training, we utilize the language modeling loss~\citep{gpt}, whereas the model predicts the next word in a sequence.

So, the overall training objective is to minimize:
\begin{equation}
    \mathcal{L} = \mathcal{L}_{\text{task}} + \alpha\mathcal{L}_{\text{balance}},
\end{equation}
where $\alpha$ is a coefficient for load balancing.

\section{Experiments}
\label{Sec:Exp}
\subsection{Experimental Setup}

\noindent\textbf{Compared Baselines.} We include two baseline models for comparison purposes: (1) \textbf{\base}, which represents a Transformer decoder without the incorporation of sparsely-activated parallel modules (i.e., SMoE layer). (2) \textbf{\xmoe}, which is our implementation based on the approach proposed by~\citet{chi2022representation}. We build our \our{} upon \xmoe{} and uses identical settings to those employed in \xmoe{}.
Note that the all these models are pre-trained using the same training data as~\our{}, and we ensure that the parameter count of our model remains consistent with or lower than that of \xmoe{}, ensuring a fair and equitable comparison. A detailed analysis and comparison about parameter and computational complexity can be found in Section~\ref{subsec: C&P} and Table~\ref{tab: paramter count}.

\noindent\textbf{Pre-training Data.} 
We detail the pre-training data of \our{}, demonstrating its effectiveness in enabling denser expert activation and finer-grained understanding through a series of experiments. These experiments are organized into three thematic categories: (1) For the English-focused experiments, we pretrain both the baseline models and \our{} on the RedPajama dataset~\cite{together2023redpajama}, which is an open-source pre-training dataset comprising sources such as Common Crawl, C4~\cite{t5}, Wikipedia, and additional curated datasets. The pretraining is conducted using GPT tasks to predict the next word in a sequence. (2) In the context of multilingual representation, we pretrain the baseline models and \our{} on the multilingual Wikipedia, following the approach described in XLM~\cite{lample2019crosslingual}, again utilizing GPT tasks. (3) For the multimodal domain, we pretrain all compared baselines and \our{} on masked multi-modality modeling task on both monomodal and multimodal data ($14$M images, $160$GB documents and $21$M image-text pairs following~\citet{wang2022BEiTv3}), and we present the details of these pre-training data in Appendix~\ref{app: pre-training data}.

\noindent\textbf{Model Architecture and Hyperparameters.} For all experiments, we use the \xmoe{}~\citet{chi2022representation} as our backbone architecture to build our \our{}, which has shown better performance than prior SMoE models such as Switch Transformers~\citep{fedus2022switch} on cross-lingual understanding benchmarks. 
For English-focused Language Modeling and Multi-lingual Language Modeling, we construct \base{}, \xmoe{} and \our{} using the Transformer~\citep{transformer} decoder (L = 12, H = 768, A = 12) with the GPT-4\footnote{\url{https://github.com/openai/tiktoken}} vocabulary as the backbone architecture. The pre-training procedure takes $14$ days on $2$ NVIDIA DGX-2 Stations. 
For Masked Multi-modal Modeling, we build \base{}, \xmoe{} and \our{} following the same Transformer encoder architecture as BEiT v3~\citep{wang2022BEiTv3}. The pre-training procedure takes $4$ days on $2$ NVIDIA DGX-2 Stations. For all three pre-training tasks, we set the head number $h$ = 4. More details about architecture and training hyperparameters can be found in Appendix~\ref{app:params:Hyperparameters} and \ref{app:params:pt}.

\subsection{Perplexity Evaluation}
%
We examined the validation perplexity curves for all pre-trained models and pre-training tasks under two expert settings~(8 experts and 32 experts). The perplexity trends are depicted in Figure~\ref{fig:ppl}, with the final perplexity values listed in Table~\ref{table:main_ppl}.
We can observe that as training progresses: 1) the perplexity of our~\our{} remained lower in comparison to the compared baselines, indicating more effective learning; 2) \our{} achieved the lowest perplexity across three distinct experimental setups; 3) an increase in the number of experts led to a corresponding decrease in the perplexity of \our{}, suggesting that the model benefits from enhanced representation learning capabilities as more experts are incorporated.
These results collectively demonstrate the superiority of \our{} in terms of learning efficiency and language representation across multiple pre-training paradigms.

\begin{table}[t]
\vspace{-2mm}
\caption{Results of upstream perplexity evaluation. We report the validation perplexity cross two setting: 8 experts and 32 experts.}
\centering
\resizebox{0.9\linewidth}{!}{
\renewcommand\tabcolsep{9.5pt}
\begin{tabular}{lcc}
\toprule
\multirow{2}{*}{\bf Model}  & \multicolumn{2}{c}{\bf Perplexity $\downarrow$} \\
\cmidrule(lr){2-3}
& 8 Experts & 32 Experts \\
\midrule
\multicolumn{3}{l}{~~\textit{English-focused language modeling}} \\
Dense (without Experts) & 16.23 & 16.23 \\
\xmoe & 14.82 & 11.96 \\
\rowcolor{gray94}\our~(Ours) & \bf 12.72 & \bf 10.28 \\
\midrule
\multicolumn{3}{l}{~~\textit{Multi-lingual language modeling}} \\
Dense (without Experts) & 8.56 & 8.56 \\
\xmoe  & 7.19 & 6.02 \\
\rowcolor{gray94}\our~(Ours) & \bf 6.26 & \bf 5.09 \\
\midrule
\multicolumn{3}{l}{~~\textit{Masked multi-modal modeling}} \\
Dense (without Experts) & 17.95 & 17.95 \\
\xmoe  & 16.34 & 12.68 \\
\rowcolor{gray94}\our~(Ours) & \bf 14.73 & \bf 10.87 \\
\bottomrule
\end{tabular}}
\label{table:main_ppl}
\vspace{-6mm}
\end{table}


\begin{table*}[t]
\caption{Accuracy / accuracy-normalization scores for language understanding tasks using the LLM Evaluation Harness~\cite{eval-harness}.}
\centering
\resizebox{\linewidth}{!}{
\renewcommand\tabcolsep{10.0pt}
\begin{tabular}{@{}lccccccccccccc}
\toprule
\bf Model & \small ARC-Challenge & \small ARC-Easy & \small RTE & \small BookQA & \small Winogrande & \small PiQA & \small BoolQ & \small HellaSwag  & \small TruthfulQA (mc1/mc2) & \small Avg \\ \midrule
\base & 18.1/23.3 & 44.9/39.7 & 51.5 & 17.1/29.0 & 48.2 & 66.6 & 55.0 & 29.7/34.1 & 24.1/39.3 & 37.2 \\
\midrule
\multicolumn{5}{l}{~\textit{Experts Number $N = 8$}} \\
\xmoe &  19.0/24.7 & 48.3/42.0 & 52.7 & 17.4/29.8 & 50.3 & 67.9 & 58.4 & 31.4/35.7 & 24.3/40.2 & 38.7 \\
\rowcolor{gray94} \our{} & \bf 19.6/25.2 & \bf 50.2/42.2 & \bf 53.0 & \bf 18.2/30.3 & \bf 51.1 & \bf 68.7 & \bf 59.6 & \bf 33.2/40.3 & \bf 24.7/\bf 40.9 & \bf 39.8\\
\midrule
\multicolumn{5}{l}{~\textit{Experts Number $N = 32$}} \\
\xmoe & 19.4/24.8 & 50.4/42.5 & 52.7 & 17.8/30.0 & 51.3 & 68.8 & 52.8 & 33.4/40.1 & 24.3/39.1 & 39.1\\
\rowcolor{gray94} \our{} & \bf 21.4/26.8 & \bf 50.6/44.8 & \bf 53.4 & \bf 18.8/31.6 & \bf 53.8 & \bf 69.3 & \bf 56.6 & \bf 35.0/42.1 & \bf 24.8/\bf 39.5 & \bf 40.6 \\
\bottomrule
\end{tabular}}
\label{table:nr}
\vspace{-6mm}
\end{table*}

\begin{table*}[t]
\begin{minipage}[ht]{0.58\linewidth}
\centering
\caption{Accuracy / accuracy-normalization scores on multilingual understanding tasks using the LLM Evaluation Harness~\cite{eval-harness}. }
\centering
\resizebox{\linewidth}{!}{
\renewcommand\tabcolsep{3.0pt}
\begin{tabular}{@{}lccccccccccccccc}
\toprule
\bf Model & bg & de & el & en & es & fr & hi & ru & sw & th & tr & ur & vi & zh & Avg \\ 
\midrule
\base & 33.1 & 33.3 & 33.0 & 35.1 & 32.8 & 34.4 & 33.6 & 34.2 & 33.3 & 33.1 & 33.3 & 33.9 & 33.5 & 32.9 & 33.5 \\
\midrule
\multicolumn{5}{l}{~\textit{\small Experts Number $N = 8$}} \\
\xmoe & 33.9 & \bf 33.4 & 33.4 & 37.3 & 33.3 & 35.9 & 34.5 & 35.0 & 33.5 & 33.6 & 33.4 & 34.2 & 33.3 & 33.2 & 34.1 \\
\rowcolor{gray94} \our{} & \bf 34.4 & 33.2 & \bf 33.9 & \bf 40.1 & \bf 34.0 & \bf 36.4 & \bf 34.6 & \bf 35.2 & \bf 33.8 & \bf 34.4 & \bf 33.3 & \bf 34.7 & \bf 34.6 & \bf 33.5 & \bf 34.7\\
\midrule
\multicolumn{5}{l}{~\textit{\small Experts Number $N = 32$}} \\
\xmoe & 34.5 & 34.5 & 33.4 & 39.6 & 33.1 & 35.3 & 34.1 & 35.4 & 33.6 & 34.7 & 33.7 & 33.6 & 34.5 & 33.3 & 34.5\\
\rowcolor{gray94} \our{} & \bf 35.8 & \bf 35.6 & \bf 34.1 & \bf 40.7 & \bf 33.9 & \bf 36.7 & \bf 34.4 & \bf 36.3 & \bf 34.3 & \bf 36.0 & \bf 34.1 & \bf 34.3 & \bf 35.2 & \bf 33.6 & \bf 35.3  \\
\bottomrule
\end{tabular}}
\label{table:mt}
\end{minipage}
\hspace{0.5pt}
\begin{minipage}[ht]{0.41\linewidth}
\centering
\caption{Results of visual question answering, visual reasoning, and image captioning tasks.}
\resizebox{\linewidth}{!}{
\renewcommand\tabcolsep{2.0pt}
\begin{tabular}{@{}lcccccccc@{}}
\toprule
\multirow{2}{*}{\bf Model} & \multicolumn{2}{c}{\small\bf VQAv2} & \multicolumn{2}{c}{\small\bf NLVR2} & \multicolumn{4}{c}{\small\bf COCO Captioning} \\
\cmidrule[0.5pt](rl){2-3} \cmidrule[0.5pt](rl){4-5} \cmidrule[0.5pt](rl){6-9}
 & test-dev & test-std & dev & test-P & B@4 & M & C & S \\
\midrule
\base & 65.9 & 66.0 & 73.8 & 74.2 & 35.9 & 29.3 & 120.5 & 19.6 \\
\midrule
\multicolumn{5}{l}{~\textit{\small Experts Number $N = 8$}} \\
\xmoe{} &  68.4 & 69.7  & 75.5 & 76.1 & 38.1 & 30.2 & 122.9 & 21.3 \\
\rowcolor{gray94}\our{} & \bf 70.1 & \bf 71.4 & \bf 77.0 & \bf 77.8 & \bf 39.7 & \bf 33.1 & \bf 124.1  & \bf 23.0 \\
\bottomrule
\end{tabular}}
\label{tbl:results:vqa_nlvr2_captioning}
\end{minipage}
\vspace{-4mm}
\end{table*}

\subsection{Downstream Evaluation}
For each pre-training task, we conduct corresponding downstream evaluation to validate the efficacy of~\our{}.

\noindent\textbf{English-focused Language Modeling.} 
We evaluate our models on a total of 9 different zero-shot benchmarks to assess their abilities across various natural language tasks like common sense reasoning, general language understanding and knowledge understanding using the LLM Evaluation Harness~\cite{eval-harness}.
As shown in Table~\ref{table:nr}, comparing \xmoe{} with the \base{} model, \xmoe{} show notable improvement, indicating that SMoE models~(e.g., \xmoe{}) benefit from the large model capacity. Overall, for all benchmarks, our \our{} obtains the best performance, achieving an average performance gain of 1.1 for 8 experts setting and 1.5 for 32 experts setting compared to \xmoe{}, demonstrating the effectiveness of our proposed multi-head mechanism on modeling English-focused language.

\noindent\textbf{Multi-lingual Language Modeling.}
We evaluate our multi-lingual language models on the cross-lingual natural language inference (XNLI) corpus~\cite{xnli}, which is the extension of the multi-genre NLI (MultiNLI) corpus to 14 languages. We follow the LLM Evaluation Harness pipeline and use the zero-shot setting to evaluate the multi-lingual ability.
%
Table~\ref{table:mt} presents the zero-shot evaluation results on XNLI task. Similarly, \xmoe{} benefit from the large model capacity and show notable improvement compared with \base{} model. Overall, \our{} obtains the best performance, surpassing \xmoe{} by an average performance gain of 0.6 for 8 experts setting and 0.8 for 32 experts setting. Comparing \our{} with the \xmoe{}, it shows that \our{} models provide consistent gains on downstream tasks, demonstrating the effectiveness of our proposed multi-head mechanism on modeling cross-lingual natural language.

\begin{table*}[t]
\centering
\begin{minipage}[ht]{0.33\textwidth}
\centering
\caption{Comparison results for different head number $h$. S-Dim denotes the dimension length of sub-tokens.}
\resizebox{\linewidth}{!}{
\renewcommand\tabcolsep{8.0pt}
\begin{tabular}{lcc|c}
\toprule
\bf Model & \bf Heads $h$ & S-Dim & \bf Perplexity \\
\midrule
\xmoe{} & - & - & 14.82\\
\midrule
\multirow{5}{*}{\our{}} & 2 & 384 & 12.87 \\
& 4 & 192 & 12.72 \\
& 6 & 128 & \bf 12.41 \\
& 8 & 96 & 12.95 \\
& 12 & 64 & 13.28 \\
\bottomrule
\end{tabular}}
\label{table:ab_heads}
\end{minipage}
\hspace{0.5pt}
\begin{minipage}[ht]{0.33\textwidth}
\centering
\caption{Ablation studies of \our{} components: MLP layers and the Token-Splitting-Merging (TSM, Eq.~\ref{EQ. sub-tokens} and Eq.~\ref{EQ: merge tokens}) operation.}
\centering
\resizebox{\linewidth}{!}{
\renewcommand\tabcolsep{10pt}
\begin{tabular}{lcc|c}
\toprule
\bf Model & MLP & TSM & \bf Perplexity \\
\midrule
Dense & \xmark & \xmark & 16.23 \\
Dense$_{w/~\text{MLP}}$ & \cmark & \xmark & 16.40 \\
\xmoe{}  & \xmark & \xmark & 14.82\\
\xmoe{}$_{w/~\text{MLP}}$ & \cmark & \xmark & 14.77 \\
\midrule
\our{}$_{w/o~\text{TS}}$ & \cmark & \xmark & 14.77 \\
\our{}$_{w/o~\text{MLP}}$ & \xmark & \cmark & 13.97 \\
\our{} & \cmark & \cmark  & \bf 12.72 \\
\bottomrule
\end{tabular}}
\label{table:components}
\end{minipage}
\hspace{0.5pt}
\begin{minipage}[ht]{0.31\textwidth}
\centering
\caption{Comparison results for different numbers of MLP layers $n$. The results are averaged over five runs.}
\centering
\resizebox{\linewidth}{!}{
\renewcommand\tabcolsep{3.5pt}
\begin{tabular}{c|ccccc}
\toprule
\multirow{2}{*}{\bf $n$} & & \bf\small Upstream & \multicolumn{3}{c}{\bf\small Downstream} \\
\cmidrule[0.5pt](rl){3-3} \cmidrule[0.5pt](rl){4-6}
& & \footnotesize Perplexity & \footnotesize RTE & \footnotesize PIQA & \footnotesize Winogrande \\
\midrule
0 & & 13.97 & 52.9 & 68.2 & 51.7 \\
1 & & 12.72 & 53.4 & \bf 69.3 & \bf 53.8 \\
2 & & \textbf{12.66} & \bf 54.0 & 68.8 & 53.3 \\
3 & & 12.87 & 53.1 & 68.8 & 52.7 \\
\bottomrule
\end{tabular}
}
\label{table:MLP_number}
\end{minipage}
\vspace{-6mm}
\end{table*}

\noindent\textbf{Masked Multi-modal Modeling.} We evaluate on the widely used vision-language understanding and generation benchmarks, including visual question answering~\citep{vqa}, visual reasoning~\citep{nlvr2} and image captioning~\citep{coco}. We report \textit{vqa-score} on VQAv2, accuracy for NLVR2. For COCO image captioning, we report BLEU@4 (B@4), METEOR (M), CIDEr (C), and SPICE (S).
Table~\ref{tbl:results:vqa_nlvr2_captioning} presents the evaluation results. For VQA task, \our{} outperforms both \base{} and \xmoe{} by a large margin, e.g., 4.24 and 1.69 points gain on test-dev split, respectively. For visual reasoning task, \our{} beats all these two baselines on both dev (1.5 points gain than \xmoe{}) and test-P split (1.7 points gain than \xmoe{}). For image captioning task, \our{} surpasses \xmoe{} by 4.2\%, 10.2\%, 9.4\% in terms of B@4, M and S, respectively.
Above results show that \xmoe{} exhibits enhanced visual information comprehension, which also validates the effectiveness of our proposed multi-head mechanism in capturing diverse semantic and detailed information within visual data.

\subsection{Ablation Studies}
\label{Sec: Ablation Studies}
This section presents experimental analysis to demonstrate the functionality of \our{}. In all comparative experiments, \emph{we ensure parameter equality across models by adjusting the internal dimensions of the experts}.

\noindent\textbf{Number of Heads $h$.} We conduct experiments by adjusting the number of heads ($h$ = $2$, $4$, $6$, $8$, and $12$) in \our{}. As shown in Table~\ref{table:ab_heads}, we find that across all settings of $h$, our model consistently outperforms the~\xmoe{}, demonstrating the effectiveness of \our{}. Besides, as the value of $h$ increases, we observe an initial improvement followed by a decline in our model's performance. This leads us to hypothesize that when $ h \leq 6$ the enhancement in performance benefits from the multi-head mechanism by activating a greater number of experts, thereby enhancing the model's effectiveness and capturing a wider range of fine-grained token information. However, as $h$ continues to increase beyond 6, the excessive subdivision of tokens may inadvertently impair their original semantic content, resulting in a decrease in model performance.

\noindent\textbf{Effect of~\our{} Components.} As shown in Figure~\ref{fig:Main_arch} (b), the multi-head mechanism utilized in our \our{} primarily incorporates two components:  the Multilayer Perceptron (MLP) layers, including the multi-head layer (Eq.~\ref{EQ: multi-head}) and merge layer (Eq.~\ref{EQ: merge layer}), and the Token-Splitting-Merging (TSM) operation (Eq.~\ref{EQ. sub-tokens} and Eq.~\ref{EQ: merge tokens}). We conduct a detailed analysis of the effectiveness of each component within our model, as well as the necessity of their integration.

\begin{figure}[t]
    \centering
    \raisebox{0.2\height}{\makebox[0.01\textwidth]{\rotatebox{90}{\centering\makecell{\small \emph{Harness}}~~~~~~~~~~\makecell{\small \emph{XNLI}}}}}
    \includegraphics[width=0.96\linewidth]{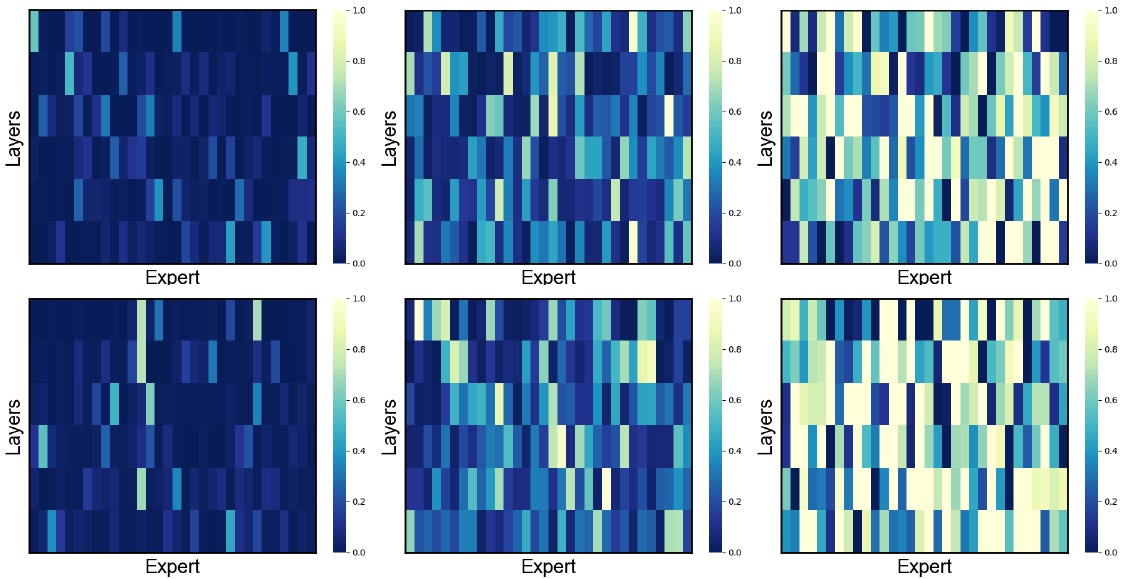}
    \\
    \vspace{-1mm}
    \hspace{1pt}
    \makebox[0.32\linewidth]{\small \xmoe{}}
    \makebox[0.32\linewidth]{\small \our{} $h = 4$}
    \makebox[0.32\linewidth]{\small \our{} $h = 8$}
    \vspace{-6mm}
    \caption{\textbf{Distribution of expert activation in \xmoe{} and \our{}} on both \emph{Harness}~\citep{eval-harness} and \emph{XNLI}~\citep{xnli} corpus, encompassing $6$ SMoE layers with $32$ experts per layer. The top of the heatmap is the first SMoE layer while the bottom is the last. Experts activation ratio is determined by calculating the ratio of each expert's selection frequency in each MoE layer to the total number of tokens.}
    \label{fig:usage}
    \vspace{-5mm}
\end{figure}

The results are presented in Table~\ref{table:components}. A comparative analysis between \base{} v.s. \base{}$_{w/o~\text{MLP}}$, as well as \xmoe{} v.s. \xmoe{}$_{w/~\text{MLP}}$, reveals that introduction of the MLP layer does not enhance the model's performance. Similarly, when comparing \our{} with \our{}$_{w/o~\text{MLP}}$, it becomes evident that the inclusion of only the MLP, in the absence of the TS, also does not yield any improvement in the model’s effectiveness. The parameter quantities of the models being compared pairwise are equal.

An intriguing observation is made when comparing \our{} with \our{}$_{w/o~\text{TS}}$. Introducing Token-Splitting-Merging (TSM) alone, without MLP, results in a slight increase in model performance. In contrast, a significant enhancement in model performance is only achieved when both MLP and TS are incorporated simultaneously. We hypothesize that introduction of TS, without the integration of MLP, activates more experts, but the segmentation and merging of the model appears overly straightforward and abrupt in its execution. This limitation hinders the model's ability to meaningfully segment tokens into sub-tokens and effectively merge the diverse information gathered from different expert spaces.

\noindent\textbf{Number of MLP layers.} We explore the impact of varying the number of layers ($n$ = 0, 1, 2, 3) in MLP on~\our{} performance. For configurations exceeding a single layer, ReLU activation functions were incorporated between MLP layers to ensure the non-linearity of transformations. The parameter quantities of the models being compared are equal.
Upon analyzing the results in Table~\ref{table:MLP_number}, we observe that increasing the number of MLP layers beyond one had a negligible impact on the model's performance. This indicates that a single-layer MLP is sufficient for accomplishing token segmentation and fusion.

\begin{figure}[t]
    \centering
    \includegraphics[width=0.47\linewidth]{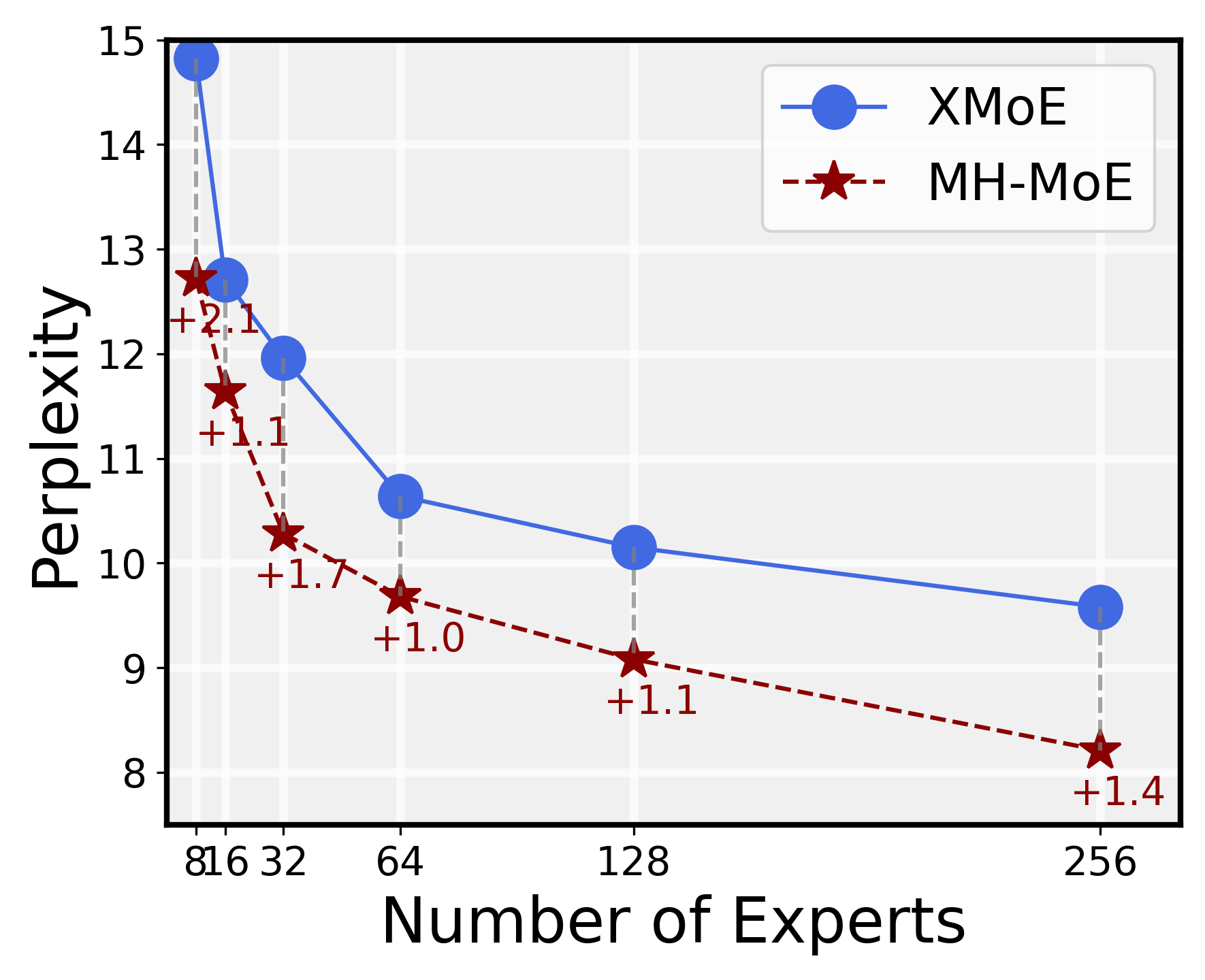}
    \hspace{1pt}
    \includegraphics[width=0.47\linewidth]{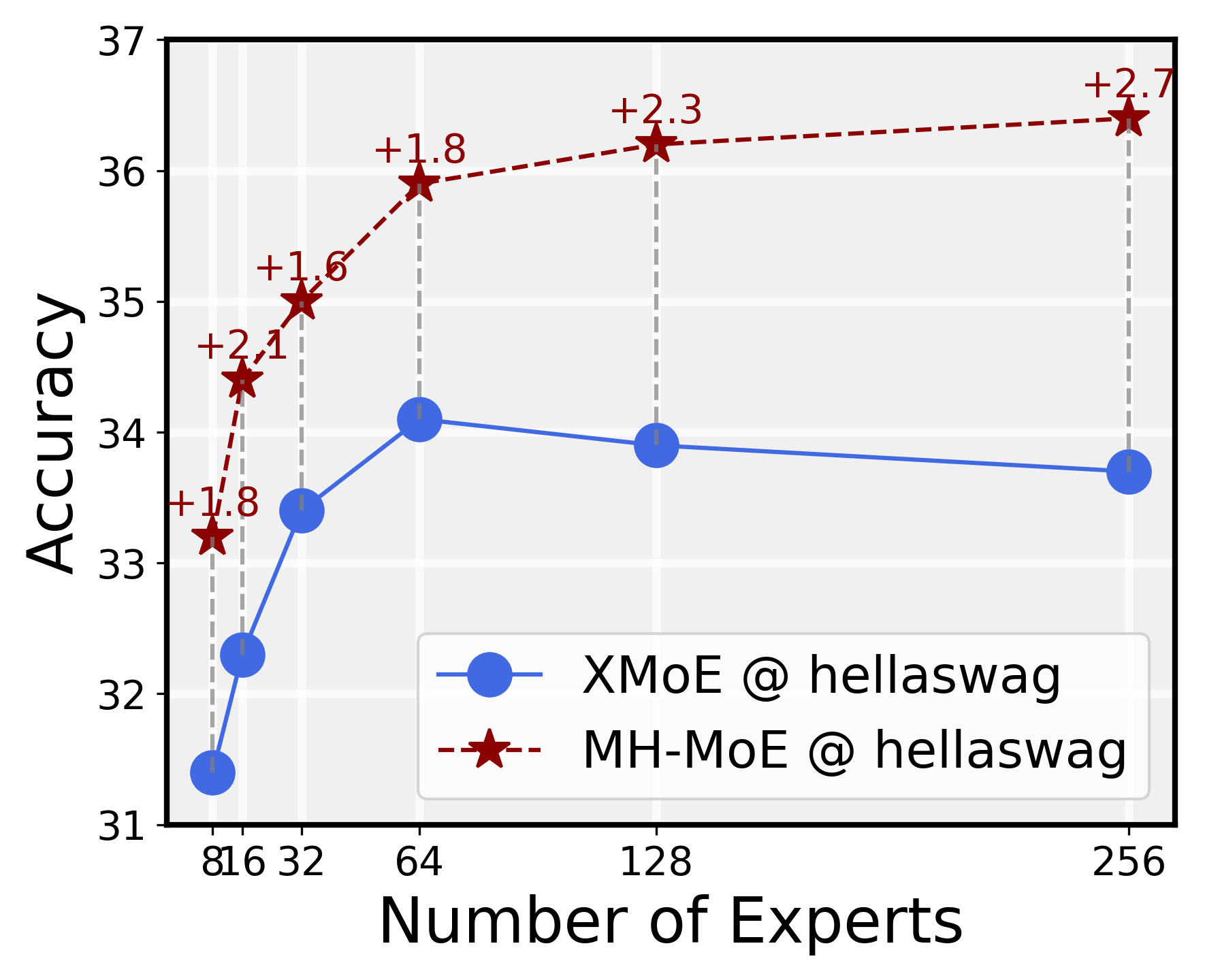}
    \\
    \vspace{-1mm}
    \hspace{1mm}
    \makebox[0.47\linewidth]{\small (a) Upstream}
    \hspace{1pt}
    \makebox[0.47\linewidth]{\small (b) Downstream}
    \vspace{-2mm}
    \caption{\textbf{Upstream and downstream results for scaling up the number of experts in \xmoe{} and \our{}}. (a) Training perplexity ($\downarrow$) when scaling the number of experts. (b) Downstream performance (accuracy scores $\uparrow$) on hellaswag when scaling the number of experts.}
    \label{fig: scale-ppl}
    \vspace{-5mm}
\end{figure}

\vspace{-2mm}
\section{Analysis}

\subsection{Experts Activation Analysis}

\noindent\textbf{Experts Activation.} We visualize the activation of each
expert varies across parallel expert layers for \xmoe{} and \our{} at Figure~\ref{fig:usage}. It can be observed that: 1) \xmoe{} demonstrate a more skewed distribution, wherein a significant portion of experts remain inactivated all the time. 2) Our \our{} achieves a denser expert activation compared to \xmoe{}, effectively mitigating the issue of low expert utilization. 3) As the number of heads $h$ increases, the expert activation frequency in \our{} also rises. 

\noindent\textbf{Scalability.}
We explore the scalability for both \xmoe{} and \our{} by scaling up the number of experts from 8 to 256~(about 7B parameters). 
\begin{figure}[t]
    \centering
    \includegraphics[width=0.95\linewidth]{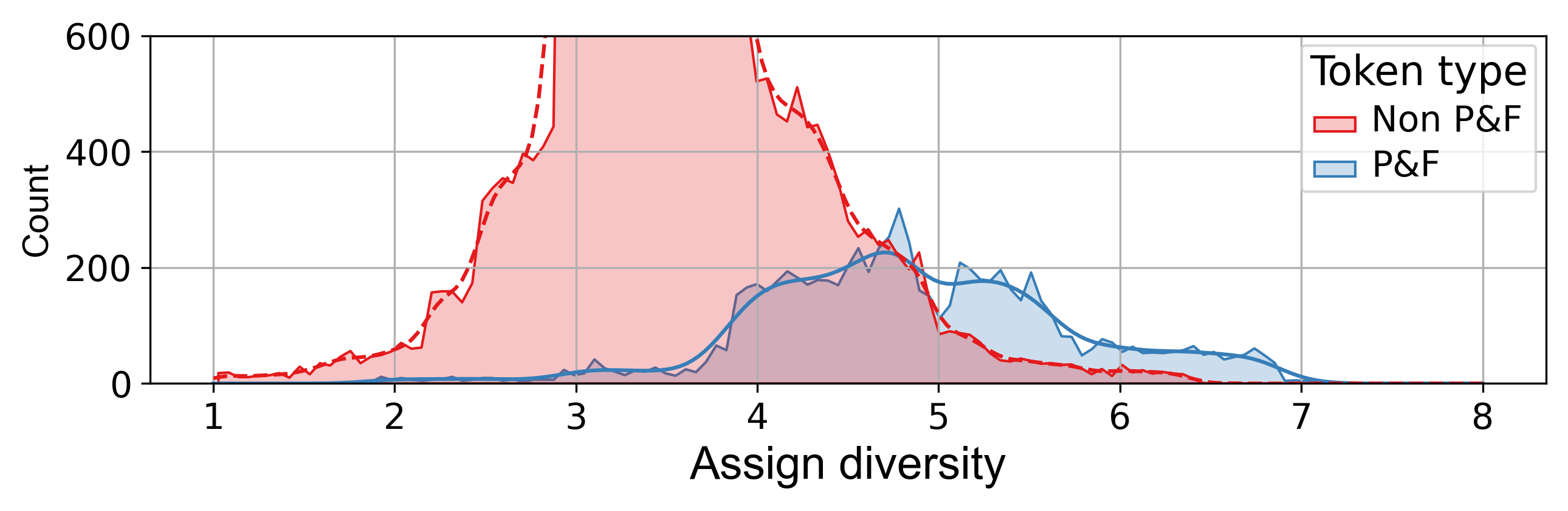}
    \vspace{-5mm}
    \caption{\textbf{Comparison for sub-tokens assign diversity} (the number of different experts they are routed to) for P\&F and Non P\&F tokens. P\&F tokens refer to the polysemous and false cognate words identified by GPT-4, while Non P\&F tokens represent the remaining words.}
    \label{fig: text-assign}
    \vspace{-7mm}
\end{figure}
For upstream performance, as shown in Figure~\ref{fig: scale-ppl} (a), with the increase of experts, our \our{} could bring more gains. It is because \our{} could mitigate the low expert activation problem effectively. With this ability, the superiority of the large-scale SMoE model will be better exerted, thereby achieving the improvement of the upper bound of SMoE with more experts. Detailed validation perplexity curves for these scaling up experiments can be found in Figure~\ref{fig:sacleup-ppl} at Appendix~\ref{app: Scale Up}.
For downstream performance shown in Figure~\ref{fig: scale-ppl} (b), for~\xmoe{}, expert number = 64 is the upper bound, meaning that continuing to increase the number of experts will not bring any gain. Our~\our{} not only has a performance advantage over the \xmoe{} with the same number of experts, but also improves the upper bound from 64 to 256, which demonstrates the effectiveness of the scalability of our~\our{} on downstream tasks.

\subsection{Fine-grained understanding Analysis}
\vspace{-1mm}
In Section~\ref{Sec:Exp}, our model excels in multiple upstream and downstream tasks, demonstrating superior fine-grained modeling capabilities, both for languages and images. In this section, we delve into a more granular analysis to validate how the multi-head mechanism aids \our{} in capturing diverse and intricate semantic information that is often challenging to comprehend, e.g., polysemous and false cognates words (denoted as PF tokens) in languages, and semantically-rich areas in images. Note that for languages data, we utilized the GPT-4 API~\citep{OpenAI2023GPT4TR} to extract polysemous words and false cognates from the XNLI~\citep{xnli} corpus, and the corresponding prompt can be found in Table~\ref{[tab: prompt]}.

\begin{figure}[t]
    \centering
    \includegraphics[width=\linewidth]{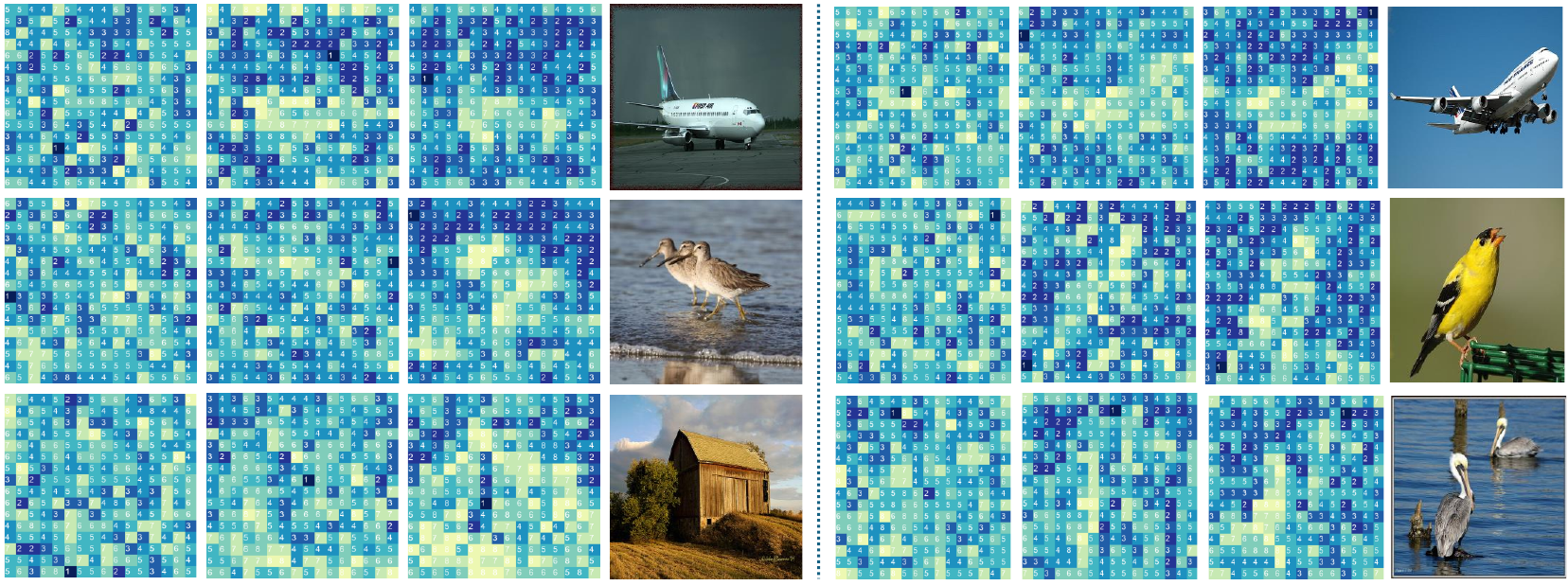}
    \\
    \vspace{-2mm}
    \makebox[0.115\linewidth]{\scriptsize 100k}
    \makebox[0.115\linewidth]{\scriptsize 200k}
    \makebox[0.115\linewidth]{\scriptsize 250k}
    \makebox[0.119\linewidth]{\scriptsize RGB}
    \hfill
    \makebox[0.115\linewidth]{\scriptsize 100k}
    \makebox[0.115\linewidth]{\scriptsize 200k}
    \makebox[0.115\linewidth]{\scriptsize 250k}
    \makebox[0.115\linewidth]{\scriptsize RGB}
    \vspace{-8mm}
    \caption{\textbf{Assign diversity of sub-tokens split from different patches} in vision data with respect to training steps (100k $\rightarrow$ 200k $\rightarrow$ 250k steps). Brighter regions indicate sub-tokens split from this patch are distributed to a greater number of diverse experts.}
    \label{fig: image-assign}
    \vspace{-4mm}
\end{figure}

\noindent\textbf{Experts Assign within Token.} 
For languages data, we compute and compare the divergence levels (i.e., the number of different experts these sub-tokens are routed to) of sub-tokens split from PF tokens and Non-PF tokens. We conduct on \our{} with 8 heads ($h$=8) and represent the divergence of each token by calculating the mean divergence across the model's various layers. The results, presented in Figure~\ref{fig: text-assign}, clearly demonstrate that the distribution of divergence for PF tokens is significantly skewed towards the right when compared to that of Non-PF tokens. This indicates that, in the \our{}'s inference process, PF tokens route their sub-tokens to a greater number of different experts, thereby capturing diverse semantic information in contrast to Non-PF tokens for a better polysemous and false cognates word modeling. 


For image data, we analyzed how the divergence levels of different patches evolve during the training process, as illustrated in Figure~\ref{fig: image-assign}. Interestingly, we observe that as the training steps increase, the divergence levels gradually increase in high-frequency texture regions (or regions with rich semantics), while the divergence levels in low-frequency texture regions gradually decrease. This indicates that during the training process, \our{} tends to route tokens from areas with complex textures to a greater variety of experts, thereby enhancing the finer-grained understanding of the semantics in that region. For more visualization examples, please refer to the Figure~\ref{fig: image-assign-app} at Appendix~\ref{app: visualization}.

\subsection{Complexity \& Parameter Analysis.}
\vspace{-1mm}
\label{subsec: C&P}
We present a analysis of Complexity \& Parameter for \xmoe{} and \our{} in Appendix~\ref{Appendix: Com & Param}, to validate that for all experiments setting, the computational and parameter cost of our \our{} are both lower than SMoE. Besides, a detailed parameter count for all experiments and comparable models can be seen in~Table~\ref{tab: paramter count}.

\vspace{-2mm}
\section{Conclusion}
In this paper, we study how we can to achieve a denser experts activation without introducing additional cost, while improving the fine-grained understanding ability. With the proposed \ourfull{}, we can easily implement the aforementioned functionality. Furthermore, the simplicity of \our{} allows it to integrate with other SMoE frameworks to enhance performance easily. Extensive empirical results across three tasks demonstrate the effectiveness of \our{}.



\section{Broader Impact}
In previous NLP pipelines, the dimension of word tokens has been conventionally maintained constant during both training and inference stages. We are the first to attempt token segmentation outside of the multi-head attention module, aiming to enhance the model's capabilities in several respects, including a more nuanced and multifaceted understanding of the token content as well as fostering a sparser network architecture. We believe this to be a counterintuitive yet worthwhile exploration in the field.

\bibliography{example_paper}
\bibliographystyle{icml2023}

\newpage
\appendix
\onecolumn


\section{Pre-training Data of Masked multi-modal modeling task}
\label{app: pre-training data}
Table~\ref{tab:pretraining_data} presents of pre-training data in Masked multi-modal modeling task. For multi-modal data, there are about $15$M images and $21$M image-text pairs collected from five public datasets: Conceptual 12M (CC12M)~\citep{cc12m}, Conceptual Captions (CC3M)~\citep{gcc}, SBU Captions (SBU)~\citep{sbu}, COCO~\citep{coco} and Visual Genome (VG)~\citep{vg}.
For monomodal data, we use $14$M images from ImageNet-21K and $160$GB text corpora~\citep{unilm2} from English Wikipedia, BookCorpus~\citep{bookcorpus}, OpenWebText\footnote{\url{http://skylion007.github.io/OpenWebTextCorpus}}, CC-News~\citep{roberta}, and Stories~\citep{stories_data}.

\begin{table*}[ht]
\vspace{-4mm}S
\centering
\caption{Pretraining data of Masked multi-modal modeling task. All the data are academically accessible.}
\small
\begin{tabular}{@{}lll@{}}
\toprule
\bf Data & \bf Source & \bf Size \\
\midrule
Image-Text Pair & CC12M, CC3M, SBU, COCO, VG & 21M pairs \\
Image & ImageNet-21K & 14M images \\
Text & English Wikipedia, BookCorpus, OpenWebText, CC-News, Stories & 160GB documents \\
\bottomrule
\end{tabular}
\label{tab:pretraining_data}
\end{table*}

\section{Model Hyperparameters of Language modeling tasks}
\label{app:params:Hyperparameters}

Table~\ref{table:mhhyperparameters} presents the model hyperparameters of \xmoe{} and \our{} for both English-focused language modeling and Multi-lingual language modeling tasks. The gating temperature $\tau_0$ is initialized as $0.3$ and $0.07$ for the softmax gating and sigmoid gating, respectively.
We use the same vocabulary as \mbox{XLM-R}~\citep{xlmr} with 250K subwords tokenized by SentencePiece~\citep{sentencepiece}.

\begin{table}[ht]
\vspace{-4mm}
\caption{Model hyperparameters of Dense, \xmoe{} and \our{}. The SMoE frequency refers to how many experts each token will be assigned to, i.e., the value of k in the Top-
 expert selection.}
\label{table:mhhyperparameters}
\centering
\small
\begin{tabular}{lrrr}
\toprule
\bf Hyperparameters & \bf Dense & \bf \xmoe{} & \bf \our{} \\ \midrule
FFNs within layer & 2 & 2 & 2\\
Expert embedding dimension & - & 16 & $16/h$ \\
Initialized gating temperature $\tau_0$ & - & 0.3 / 0.07 & 0.3 / 0.07 \\
\midrule
Transformer blocks & 12 & 12 & 12 \\
Hidden size & 768 & 768 & 768\\
FFN inner hidden size & 3,072 & 3,072 & 3,072 $\times \beta$ \\
Attention heads & 12 & 12 & 12\\
SMoE frequency & - & 2 & 2\\
\bottomrule
\end{tabular}
\end{table}

\section{Hyperparameters for Pre-training}
\label{app:params:pt}
Table~\ref{table:pthparam} presents the hyperparameters for pre-training across three tasks: Language modeling tasks (English-focused language modeling and Multi-lingual language modeling tasks) and Masked multi-modal modeling task.

\begin{table}[ht]
\vspace{-4mm}
\caption{Pre-training hyperparameters for Language modeling tasks (English-focused language modeling and Multi-lingual language modeling tasks) and Masked multi-modal modeling task tasks.}
\label{table:pthparam}
\centering
\small
\begin{tabular}{lrr}
\toprule
\bf Hyperparameters & \bf Language modeling tasks & \bf Multi-modality modeling task\\ \midrule
Batch size & 256 & 512 \\
Optimizer & Adam & AdamW\\
Batch tokens per task & 1M & - \\
Adam $\epsilon$ & 1e-6 &1e-6 \\
Adam $\beta$ & (0.9, 0.98) & (0.9, 0.98) \\
Maximum learning rate & 5e-4 & 2.8e-3\\
Learning rate schedule & Linear decay &  Cosine decay\\
Warmup steps & 10,000 & 10,000\\
Weight decay & 0.01 & 0.05\\
Transformer dropout & 0.1 & 0.1 \\
Dropout & 0 & 0\\
Load balancing coefficient & 1e-2  & 1e-2 \\
\bottomrule
\end{tabular}
\end{table}

\begin{table}[ht]
\vspace{-4mm}
\caption{Parameter count setting of \xmoe{} and \our{} in our experiments for English-focused language modeling, Multi-lingual language modeling and Masked multi-modality modeling tasks. ``non-expert param'' refers to the parameters that are not part of the expert networks, such as the attention layer, router, etc., while ``expert params'' represents the total number of parameters in the parallel expert networks. For Dense models, since there are no expert network layers, we only list the total number of parameters. All models under the same task utilize the same architecture and hyperparameters, following identical training settings and steps.}
\centering
\centering
\resizebox{\linewidth}{!}{
\renewcommand\tabcolsep{8.0pt}
\begin{tabular}{lrrrrrrr}
\toprule
\multirow{2}{*}{\bf Expert Setting} & \multicolumn{1}{c}{\bf \base{}} & \multicolumn{3}{c}{\bf \xmoe{}} & \multicolumn{3}{c}{\bf \our{}}\\
\cmidrule(lrrr){2-2} \cmidrule(lrrr){3-5} \cmidrule(lrrr){6-8}
& \bf Sum & non-expert params & expert params & \bf Sum & non-expert params & expert params & \bf Sum\\ 
\midrule
\multicolumn{2}{l}{~~\textit{English-focused language modeling}} \\
0 expert & 162M & - & - & - & - & - & - \\
8 experts & - & 134M & 227M & 361M &  141M & 213M  & 354M \\
16 experts & - & 134M & 454M & 588M & 141M & 430M & 571M \\
32 experts & - & 134M & 908M & 1,042M &  141M & 898M  & 1,039M\\
64 experts & - & 134M & 1,815M & 1,949M & 141M & 1,797M & 1,938M \\
128 experts & - & 134M & 3,631M & 3,765M & 141M & 3,624M & 3,765M\\
256 experts & - & 134M & 7,263M & 7,397M & 141M & 7,230M & 7,371M \\
\midrule
\multicolumn{2}{l}{~~\textit{Multi-lingual language modeling}} \\
0 expert & 162M & - & - & - & - & - & - \\
8 experts & - & 134M & 227M & 361M & 141M & 213M & 354M \\
32 experts & - & 134M & 908M & 1,042M & 141M & 898M & 1,039M\\
\midrule
\multicolumn{2}{l}{~~\textit{Masked multi-modality modeling}} \\
0 expert & 277M & - & - & - & - & - & -\\
8 experts & - & 191M & 323M & 514M & 195M & 310M & 505M \\
32 experts & - & 191M & 1,037M & 1,228M & 195M & 1,014M & 1,209M \\
\bottomrule
\end{tabular}}
\label{tab: paramter count}
\end{table}

\clearpage
\section{Complexity \& Parameter Analysis}
\label{Appendix: Com & Param}
\subsection{Complexity}
We analysis the computational cost of~\our{}. Without loss of generality, we consider one transformer block with a single-layer SMoE containing $N$ experts, but this can be easily generalized. 

The computational cost of SMoE layer comes from two parts: 1) the projection of router. 2) linear projections by parallel experts (FFNs which contains two connected linear layers with inner hidden size $4 d$). For a input sequence $\mathbf{X} \in \mathbb{R}^{l \times d}$, the computational cost for each part is
\begin{align}
    \Theta_{\text{router}} &= l \times d \times N = l d N,\\
    \Theta_{\text{experts}} &= l \times \left(d \times 4 d + 4 d \times d\right) = 8ld^2.
\end{align}
respectively. Thus, the total computational cost of SMoE is $\Theta_{\text{SMoE}} = ld (N + 8d)$. 

For~\our{} layer, in addition to the two computations mentioned above, it also encompasses two additional distinct computational aspects: 3) projection of multi-head layer. 4) projection of merge layer. The computational cost for each part is
\begin{align}
    \Theta_{\text{multi-head}} &= l \times d \times d = ld^2, \\
    \Theta_{\text{router}} &= h \times  l \times \frac{d}{h} \times N = ldN,\\
    \Theta_{\text{experts}} &= h \times l \times \left(\frac{d}{h} \times 4 \beta d + 4 \beta d \times \frac{d}{h}\right) = 8\beta ld^2,\\
    \Theta_{\text{merge}} &= l \times d \times d = ld^2,
\end{align}
where $\beta$ is a hyperparameter employed to scale the inner hidden dimension of FFNs. In our empirical experiments, we meticulously adjust $\beta$ to ensure a parametric equilibrium between our \our{} and SMoE. So, the overall computational cost of \our{} is $\Theta_{\text{\our{}}} = ld(N + 8\beta d + \frac{N}{h})$. 

Thus we compare the computational cost between SMoE and \our{} as $\delta$:
\begin{equation}
   \delta = \Theta_{\text{SMoE}} - \Theta_{\text{\our{}}} = l d \left[ \underbrace{8d ( 1 - \beta ) - \frac{N}{h}}_{\epsilon}\right],
\end{equation}

In all of our experimental setups, the smallest $\beta$ is $\frac{63}{64}$. Thus $\epsilon$ exists an lower bound when $N = 256$ (set to the largest number of experts), $h = 4$ (set to the fewest head number) and $\beta = \frac{63}{64}$ (set to the minimum $\beta$). In considering this situation, we have $\epsilon = 8\times 768 \times \frac{1}{64} - \frac{256}{4} = 96 - 64 > 0$.
%
%
\textbf{So we validate that for all experiments setting, we have $\Theta_{\text{SMoE}} - \Theta_{\text{\our{}}} > 0$, i.e., the computational cost of our \our{} is fewer than SMoE.}

\subsection{Parameter}
For SMoE, the parameter contains two parts: 1) the parameter of router. 2) the parameter of parallel experts:
\begin{align}
    \Gamma_{\text{router}} &= d \times N, \\
    \Gamma_{\text{experts}} &= N \times \left(d\times 4d + 4d \times d\right) = 8d^2 N
\end{align}
Thus, the total parameter of SMoE is $\Gamma_{\text{SMoE}} = dN (1+8d)$.

For~\our{}, in addition to the two parts of parameter mentioned above, it also encompasses two additional aspects: 3) the parameter of multi-head layer. 4) the parameter of merge layer, while the parameter for each part is
\begin{align}
    \Gamma_{\text{head}} &= d \times d, \\
    \Gamma_{\text{router}} &= \frac{d}{h} \times N, \\
    \Gamma_{\text{experts}} &= N \times \left(\frac{d}{h}\times 4\beta d + 4\beta d \times \frac{d}{h}\right) = \frac{8\beta d^2N}{h} \\
    \Gamma_{\text{merge}} &= d \times d, \\
\end{align}
So, the overall parameter of \our{} is $\Gamma_{\text{\our{}}} = 2d^2 + \frac{dN}{h} + \frac{8\beta d^2N}{h}$. \textbf{Detailed parameter comparison can be found in Tabl~\ref{tab: paramter count}, we ensure that the parameter count of our \our{} remains consistent with or lower than that of \xmoe{}, ensuring a fair and equitable comparison all experiments.}

\clearpage
\section{PyTorch-style Code}
\label{app: Algorithm}
We also provide the PyTorch-style code in Algorithm~\ref{algorithmicxxx} to explain our \our{}, which including two main aspects: 1) \texttt{Stage 1.} The creation and initialization of multi-head layer and merge layer. 2) \texttt{Stage 2.} The segmentation of tokens, followed by processing through an expert network, and ultimately merging.

\begin{algorithm}[]
\caption{The Overall Procedures of \our{} in a PyTorch-like style.}
\label{algorithmicxxx}
\definecolor{codeblue}{rgb}{0.25,0.5,0.5}
\lstset{
  backgroundcolor=\color{white},
  basicstyle=\fontsize{9pt}{9pt}\ttfamily\selectfont,
  columns=fullflexible,
  breaklines=true,
  captionpos=b,
  commentstyle=\fontsize{9pt}{9pt}\color{codeblue},
  keywordstyle=\fontsize{9pt}{9pt},
}
\textbf{Input:} A \our{} model with L parallel SMoE layers M, the number of the experts $k$.

\begin{lstlisting}[language=Python]
# Stage 1: Initial parameter of multi-head layer & merge layer

for i in range(1, L):
    M[i].multi_head_layer = nn.Linear(hidden_dim, hidden_dim)
    M[i].merge_layer = nn.Linear(hidden_dim, hidden_dim)

     # Initialization
    nn.init.xavier_uniform_(M[i].multi_head_layer.weight, gain=1 / math.sqrt(2))
    nn.init.xavier_uniform_(M[i].merge_layer.weight)
    nn.init.constant_(M[i].merge_layer.bias, 0.0)
    
# Stage 2: The segmentation and merge of tokens for the i-th MH-MoE layer

def MHMoE_Layer(x):
    '''
    Input:
        x : Tensor shape: (batch_size, Length, hidden_dim)
        mask : Tensor shape: (batch_size, Length)

    Output:
        o : Tensor shape: (batch_size, Length, hidden_dim)
        
    heads: head number of multi_head layer
    '''

    # Processed by multi-head layer
    x = M[i].multi_head_layer(x)

    # Split token & rearrange sub-tokens in parallel
    x = x.reshape(batch_size * Length * heads, hidden_dim // heads).contiguous()
    mask = mask.reshape(-1, 1).repeat(1, heads).reshape(batch_size * Length * heads)

    # Standrad SMoE routing block
    x, mask  = router(x, mask)

    # Merge back to the original token form
    x = x.reshape(batch_size * Length, heads, dim // heads).reshape(batch_size * Length, dim).contiguous()
    o = M[i].merge_layer(x)
    
    return o
    
\end{lstlisting}
\end{algorithm}

\clearpage

\begin{figure*}[t]
    \centering
    \makeatletter\def\@captype{table}\makeatother\caption{Prompt template for identifying polysemous and false cognates in different languages.}
    \hspace{5mm}
    \includegraphics[width=0.93\textwidth]{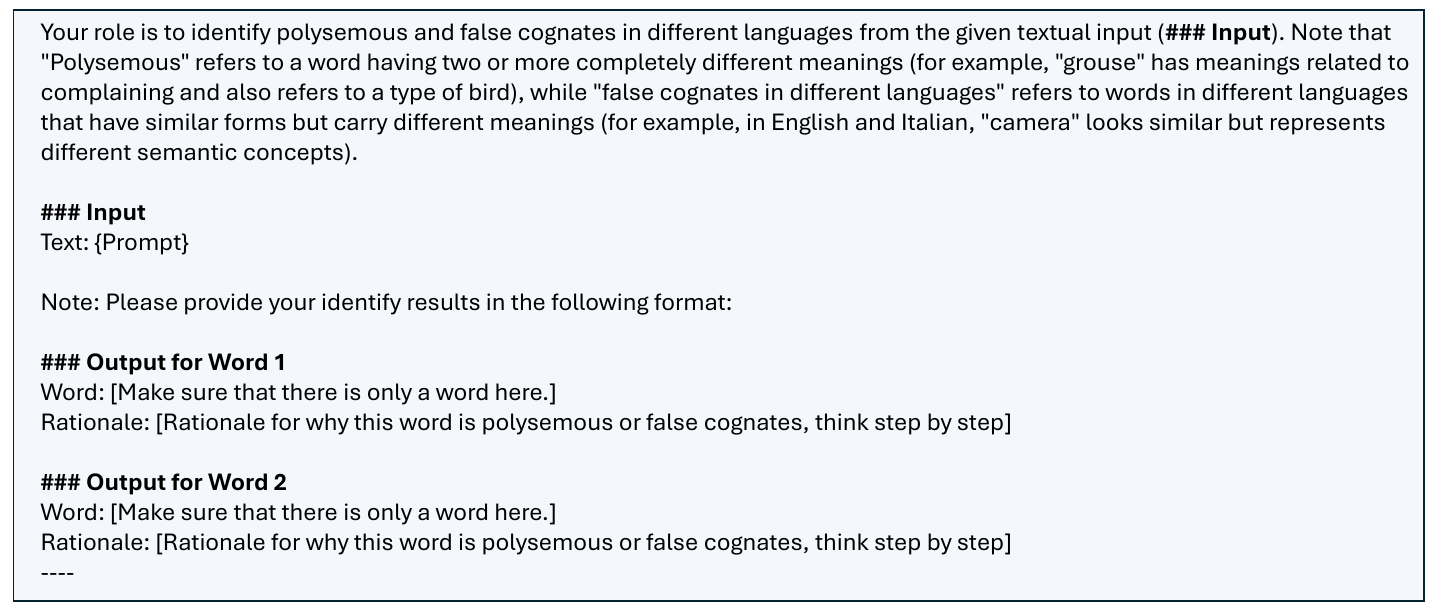}
    \label{[tab: prompt]}
    \vspace{-4mm}
\end{figure*}

\section{Visualization of training perplexity}
\label{app: Scale Up}
\begin{figure*}[t]
    \centering
    \includegraphics[width=0.95\textwidth]{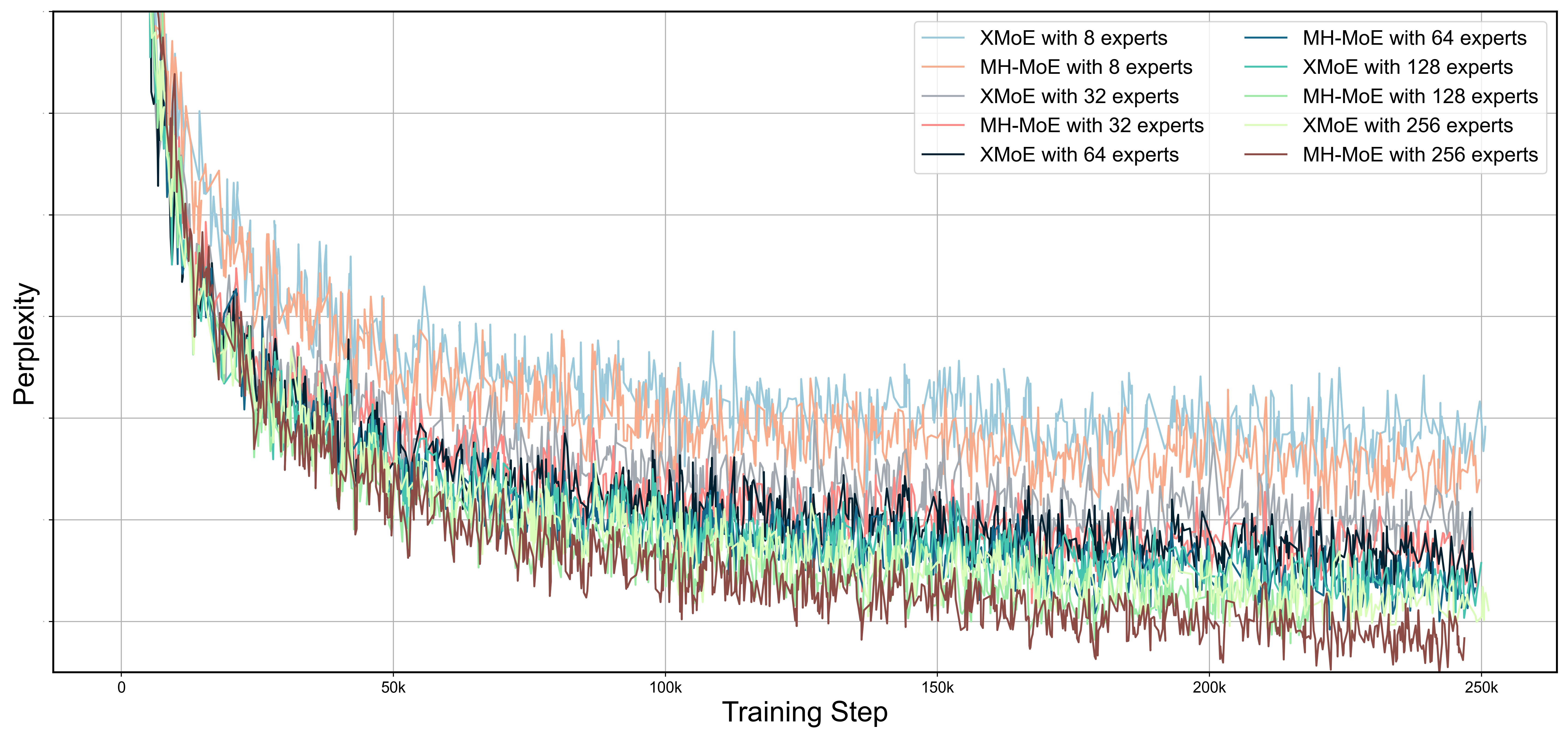}
    \vspace{-4mm}
    \caption{Validation perplexity reported for both~\xmoe{} and \our{}.}
    \label{fig:sacleup-ppl}
    \vspace{-4mm}
\end{figure*}

We provide the training perplexity curve for model training in the experimental setting of increasing the number of experts (from 8 to 256) in Figure~\ref{fig:sacleup-ppl}.

\section{Visualization}
\label{app: visualization}
\begin{figure}[t]
    \vspace{-4mm}
    \centering
    \includegraphics[width=\linewidth]{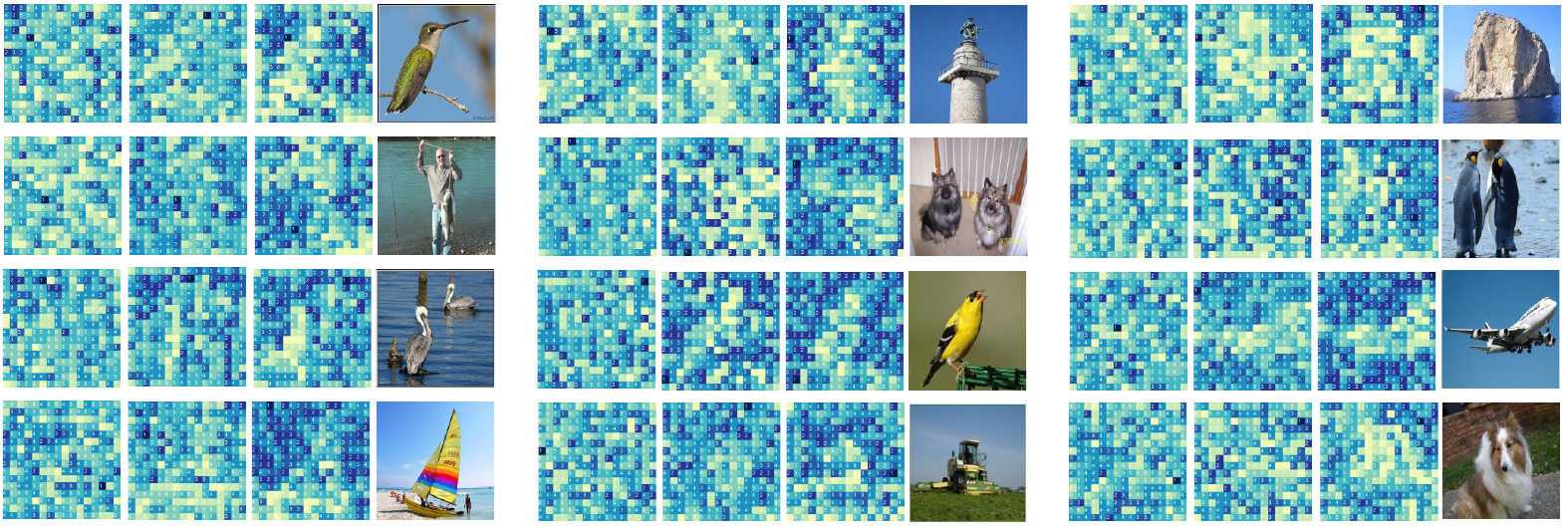}
    \\
    \vspace{-1mm}
    \makebox[0.078\linewidth]{\footnotesize 100k steps}
    \makebox[0.078\linewidth]{\footnotesize 200k steps}
    \makebox[0.078\linewidth]{\footnotesize 250k steps}
    \makebox[0.078\linewidth]{\footnotesize Input}
    \makebox[0.078\linewidth]{\footnotesize 100k steps}
    \makebox[0.078\linewidth]{\footnotesize 200k steps}
    \makebox[0.078\linewidth]{\footnotesize 250k steps}
    \makebox[0.078\linewidth]{\footnotesize Input}
    \makebox[0.078\linewidth]{\footnotesize 100k steps}
    \makebox[0.078\linewidth]{\footnotesize 200k steps}
    \makebox[0.078\linewidth]{\footnotesize 250k steps}
    \makebox[0.078\linewidth]{\footnotesize Input}
    \vspace{1mm}
    \includegraphics[width=\linewidth]{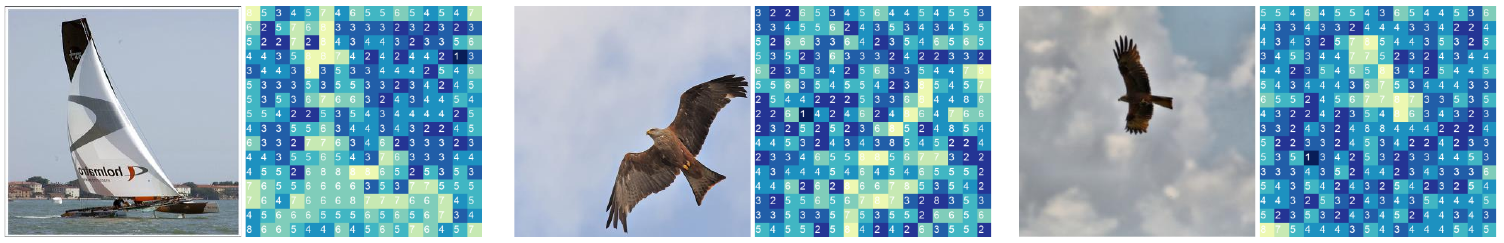}
    \includegraphics[width=\linewidth]{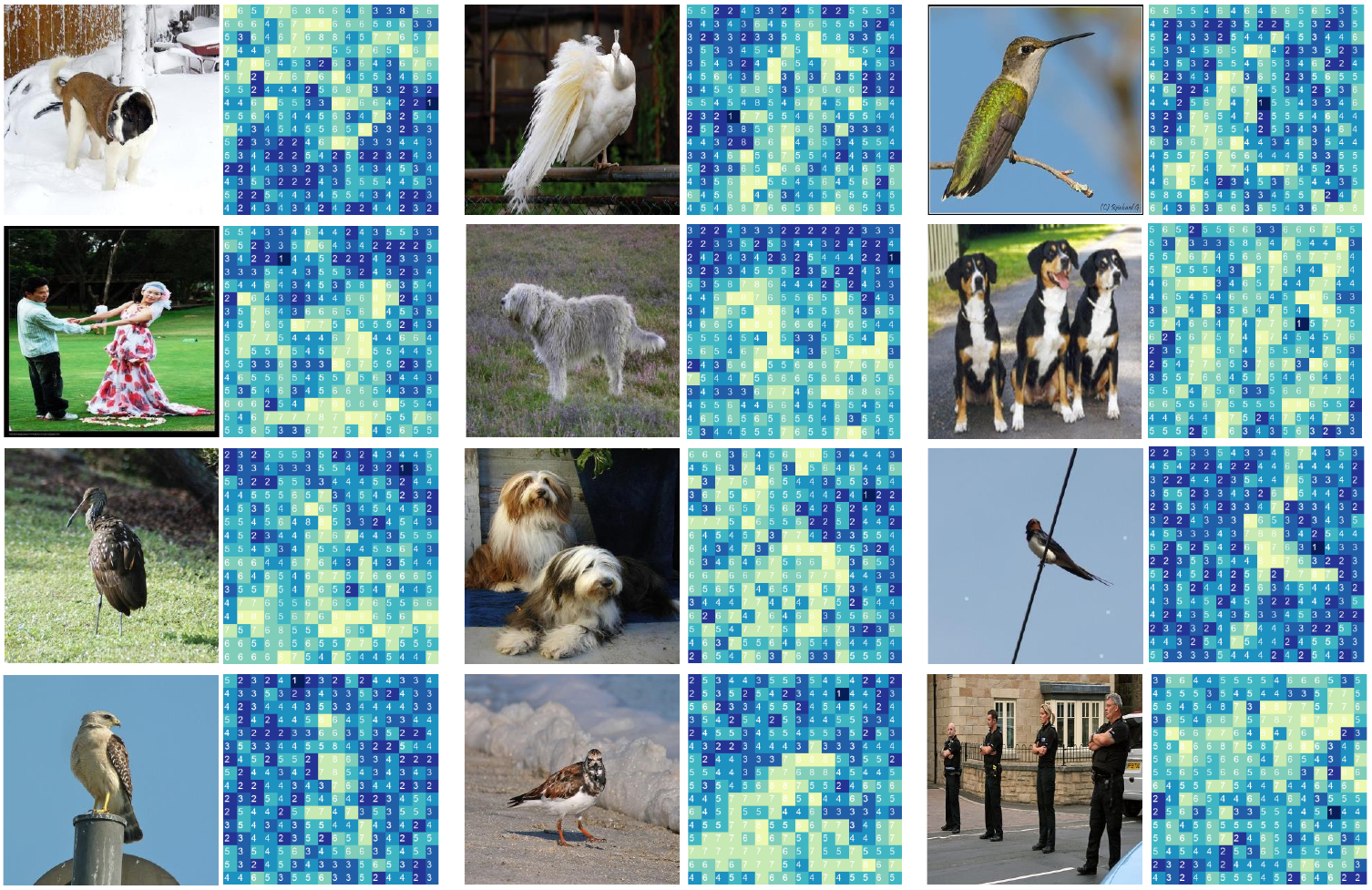}
    \vspace{-6mm}
    \caption{More visualization examples for assign diversity of sub-tokens split from different patches with respect to training steps. We analyze \our{} with 8 heads ($h$=8) as the subject. Brighter regions indicate that sub-tokens from this patch are distributed to a greater number of diverse experts, while darker regions indicate that sub-tokens are assigned to more of the same experts.}
    \label{fig: image-assign-app}
    \vspace{-4mm}
\end{figure}

We provide more visualization of variation in assign diversity for sub-tokens split from different patches in vision data at Figure~\ref{fig: image-assign-app}.

\clearpage

\end{document}